\documentclass[letterpaper,11pt]{amsart}

\usepackage{enumitem}
\usepackage{caption}
\usepackage[colorlinks]{hyperref}
\usepackage[utf8]{inputenc}
\usepackage{amssymb}
\usepackage{amsthm}
\usepackage{array}
\usepackage{multirow}
\usepackage{mathtools}
\usepackage{mathrsfs}
\usepackage{color,graphicx}
\usepackage{thmtools, thm-restate}
\usepackage{algorithm,algpseudocode}
\usepackage{etoolbox,mathabx}
\usepackage[letterpaper, margin=1.25in]{geometry}

\makeatletter
\usepackage{mathtools}

\newcommand{\cO}{\mathcal{O}}

\newcommand{\cR}{\mathcal{R}}

\newcommand{\cU}{\mathcal{U}}

\newcommand{\bitm}{\begin{itemize}}
\newcommand{\eitm}{\end{itemize}}
\newcommand{\bitme}{\begin{enumerate}[label=(\roman*),leftmargin=0.25in]}
\newcommand{\eitme}{\end{enumerate}}
\newcommand{\beq}{\begin{equation}}
\newcommand{\eeq}{\end{equation}}
\newcommand{\btcb}{\begin{tcolorbox}}
\newcommand{\etcb}{\end{tcolorbox}}
\def\bals#1\eals{\begin{align*} #1 \end{align*}}
\def\bal#1\eal{\begin{align} #1 \end{align}}

\newcommand{\Ffor}{\quad \text{ for }}
\newcommand{\Aand}{\quad \text{ and } \quad }

\newcommand\Dom\Omega

\newcommand\EE{\mathbb{E}}

\newcommand\RR{\mathbb{R}}

\newcommand\Lap\Delta

\newcommand\abs[1]{\left\lvert #1 \right\rvert}

\def\bpde#1\epde{\[\left\{\begin{aligned}#1\end{aligned}\right. \]}
\def\inbpde#1\inepde{\left\{\begin{aligned}#1\end{aligned}\right.}
\def\binpde#1\einpde{\left\{\begin{aligned}#1\end{aligned}\right.}

\newcommand\Norm[2]{\lVert { #1 } \rVert_{#2}}

\def\cD{\mathcal{D}}

\def\cL{\mathcal{L}}
\def\cU{\mathcal{U}}

\def\cR{\mathcal{R}}
\def\half{\frac{1}{2}}

\def\cU{\mathcal{U}}

\def\b0{\mathbf{0}}

\def\eps{\varepsilon}

\def\bbmat{\begin{bmatrix}[r]}
\def\ebmat{\end{bmatrix}}

\newcommand{\barr}{\begin{array}}
\newcommand{\ea}{\end{array}}
\newcommand{\bea}{\begin{eqnarray}}
\newcommand{\eea}{\end{eqnarray}}
\newcommand{\bt}{\begin{table}}
\newcommand{\et}{\end{table}}

\DeclareMathOperator\Span{span}

\DeclareMathOperator\argmin{argmin}

\DeclareMathOperator\softargmax{softargmax}
\DeclareMathOperator\row{row}
\DeclareMathOperator\col{col}

\theoremstyle{plain}

\newtheorem{genericthm}{GENERIC THEOREM ENVIRONMENT}[section]

\newtheorem{defi}[genericthm]{Definition}

\numberwithin{equation}{section}

\newcommand\tturl[1]{{\tt \scriptsize [\url{{#1}}]}}

\newcommand\ninput{n_{\text{in}}}

\newcommand\noutput{n_{\text{out}}}
\newcommand\nell{{n_{\ell}}}

\newcommand\z{\phantom{0}}

\makeatother

\ifpdf
  \DeclareGraphicsExtensions{.pdf, .jpg, .tif}
\else
  \DeclareGraphicsExtensions{.eps, .jpg}
\fi
\author{Tiana C. Johnson}
\address{Department of Mathematics, %
    Washington University in St. Louis, St. Louis, MO 63130} %
\email{j.tiana@wustl.edu}%

\author{Donsub Rim}
\address{Department of Mathematics, %
    Washington University in St. Louis, St. Louis, MO 63130} %
\email{rim@wustl.edu}%

\title[Adversarial Training Without Input Gradients]{Adversarial Training Without Input Gradients via Low-Rank Householder Expansions}

\date{\today}

\begin{document}

\begin{abstract}
This work concerns adversarial training against the small-norm adversarial
examples that arise from the inherent input instability of a trained deep neural
network. Examples in this class are small as measured in the relative
$\ell^2$-norm, and therefore lie in the neighborhood of the input on which the
model acts approximately linearly, the regime in which the perturbation
remains imperceptible. We first show that
such examples can be computed directly from the trained network parameters,
without input gradient iterations, by means of a linearization called the
low-rank Householder expansion (LRHE). The expansion describes the composed
affine map rather than any individual layer, and the directions it identifies
are read from the activation pattern already available in the forward pass. We
then propose a simple adversarial training scheme built on this construction. No
differentiation with respect to the input is performed at any point: training
requires only additional forward evaluations, with weight parameters updated by
the standard backward pass, and the inner maximization of the usual min-max
formulation is eliminated entirely. That such a regularizer exists is our main
finding: the methods that dispense with the inner search all obtain their local
geometry by differentiating with respect to the input, and we show this is not
necessary. The regularizer costs the equivalent of $2.8$ PGD steps per epoch, an
$8.7\times$ reduction relative to 40-step adversarial training on MNIST and
below the cost of 3-step training. The resulting models match three-step PGD
adversarial training for relative $\ell^2$ budgets $\eps \leq 0.02$ and 40-step
training for $\eps \leq 0.012$, falling away beyond, consistent with the
locality of the expansion.
\end{abstract}

\maketitle

\section{Introduction}

The observation that deep neural networks can be made to misclassify inputs
altered by perturbations too small to matter to a human observer is now more
than a decade old \cite{szegedy2014intriguing, goodfellow2015explaining}, and it
has proved remarkably resistant to resolution. The intervening years have
produced a large number of proposed defenses and an almost equally large number
of demonstrations that they do not work. A survey of defenses presented
at a single conference \cite{athalye2018obfuscated} found that the majority
relied on obfuscated gradients and could be circumvented once that reliance was
recognized, and similar conclusions were reached for later defenses
\cite{tramer2020adaptive}. What has resisted this attrition is adversarial
training in the min-max formulation of \cite{madry2018towards}, which augments
each parameter update with an inner search for the worst-case perturbation. It
has consistently withstood adaptive evaluation, and is the standard against
which subsequent proposals are measured
\cite{carlini2019evaluating, croce2020reliable}. The persistence of the problem
is therefore not a matter of an effective defense being unknown, but of
its cost.

That cost has a specific and unvarying structure. The inner maximization admits
no closed form and is approximated by $K$ steps of projected gradient ascent,
each requiring a gradient of the loss with respect to the input, so that
adversarial training is roughly $K$ times more expensive than ordinary training.
Efforts to reduce this expense fall into three groups. The first reuses
computation across updates: perturbations may be carried between minibatches so
that the adversarial and parameter updates share a backward pass
\cite{shafahi2019free}, or the inner loop confined to the first layer through an
optimal-control formulation \cite{zhang2019you}. The second reduces $K$
directly, usually to one, and then repairs the resulting instability; FGSM with
random initialization and a cyclic learning rate has been shown to suffice
\cite{wong2020fast}, and the catastrophic overfitting that single-step methods
exhibit can be prevented by penalizing gradient misalignment
\cite{andriushchenko2020understanding}. The third dispenses with the inner
maximization altogether, replacing it with a penalty on the local geometry of
the loss: input-gradient regularization \cite{ross2018improving}, Jacobian
regularization \cite{hoffman2019robust}, curvature regularization
\cite{moosavi2019robustness}, and local linearity regularization
\cite{qin2019adversarial}, the last of which explicitly trades a geometric
penalty for a reduction in the number of projected gradient descent (PGD) steps.
What unites these otherwise dissimilar approaches is that every one of them
obtains its information about the network's local behavior by differentiating
with respect to the input. They differ in how many such derivatives they take,
not in whether they take them.

We show that for perturbations of small relative $\ell^2$ norm this is
unnecessary. Our starting point is the low-rank Householder expansion (LRHE)
introduced in \cite{lrhe2024}. A ReLU network is piecewise affine. On the region
of input space where the activation pattern is constant, the network acts as a
single affine map. Concretely, one may write a feedforward ReLU neural network,
with the bias terms omitted, as multiplication by the matrix
\begin{equation}
  F = W_L \Sigma_{L-1} W_{L-1} \cdots \Sigma_1 W_1 ,
  \label{eq:affinemap}
\end{equation}
where each $\Sigma_\ell$ is the diagonal $0$--$1$ matrix
recording which units at layer $\ell$ are active, and $W_\ell$ are weight
matrices.  A change of activation pattern perturbs the network's local action by
a term of rank one. To see this, fix a layer $\ell$ and split the product
\eqref{eq:affinemap} on either side of $\Sigma_\ell$,
\begin{equation}
  F = C_\ell \Sigma_\ell B_\ell,
  \quad
  C_\ell = W_L \Sigma_{L-1} \cdots W_{\ell+1},
  \quad
  B_\ell = W_\ell \Sigma_{\ell-1} \cdots \Sigma_1 W_1 ,
  \label{eq:split}
\end{equation}
so that $B_\ell$ collects the factors before layer $\ell$ and $C_\ell$ those
after.  Suppose the input is displaced across a single activation boundary, so
that unit $i$ of layer $\ell$ changes from active to inactive while every other
unit retains its state. Then $\Sigma_\ell$ is replaced by $\Sigma_\ell - e_i
e_i^\top$ where $e_i$ is the $i$-th standard basis vector, and the matrix
representation of the network on inputs in the new region is
\begin{equation}
  F_1 = C_\ell (\Sigma_\ell - e_i e_i^\top) B_\ell
       = F - ( C_\ell e_i )( B_\ell^\top e_i )^{\!\top} .
  \label{eq:rankone}
\end{equation}
The perturbation is thus rank one, with left factor $C_\ell e_i$ and right
factor $B_\ell^\top e_i$, i.e. the $i$-th column of $C_\ell$ and the $i$-th row
of $B_\ell$, respectively.

The right factors $B^\top_\ell e_i$ appearing in \eqref{eq:rankone} are vectors
in input space, and it is these that determine the directions in which the
composed map is sensitive. Note that they are available from the forward pass:
$B_\ell$ is the partial composition below layer $\ell$, and $e_i$ selects the
unit in question. Through \eqref{eq:rankone} the effect of activating one unit
on the output becomes clear, and no differentiation with respect to the input is
involved at any stage.

The LRHE is derived by exploiting this property further. Substitute
$\Sigma_\ell$ as a rank-one perturbation of the identity in the form $I - v_\ell
v_\ell^\top$, in which $v_\ell$ is a unit vector. Key to this expansion is how
$v_\ell$ is found by rewriting the ReLU activation as a Householder reflection.
The reflection vectors are not learned and they are computed during the forward
pass, and require no differentiation. The expansion describes the effect of
perturbing the hidden states on the composed maps $B_\ell$ or $C_\ell$ rather
than any individual layer, so the directions it identifies are those in which
the network as a whole amplifies, not those in which an individual weight matrix
$W_\ell$ does. It was shown in \cite{lrhe2024} that adversarial examples found
for a tsunami waveform prediction task \cite{rim2022tsunami} have significant
components along these basis directions.

The present work takes that construction as given and asks a different question:
whether the subspace it identifies is the one in which adversarial perturbations
reside for a classification task, and whether restricting attention to that
subspace suffices to train robust models. Two consequences follow. First,
adversarial examples in this small-norm class can be constructed directly from
the trained parameters, with no input gradient iterations at all. Second, and
the subject of the remainder of this work, adversarial training against them
reduces to a minimization: the inner search disappears, the network is penalized
for its response along a set of directions recomputed at each forward pass, and
no derivative with respect to the input is taken at any point during training.

Since LRHE enables the identification of a linear subspace of adversarial
perturbation, this can be exploited to reduce the cost of adversarial training
by restricting the inner maximization.  Our method thus belongs to the third
group discussed above, but differs in where its geometric information
originates. Curvature, Jacobian, and local-linearity penalties are all measured,
each requiring derivative evaluations beyond those of standard training, so that
the saving relative to multi-step PGD is real but partial. The expansion instead
supplies the relevant subspace algebraically. That such a restriction should be
possible at all is suggested by an early observation \cite{tramer2017space} that
adversarial perturbations occupy a contiguous subspace whose dimension is far
below that of the input, on the order of tens at MNIST scale. If the
adversarially relevant directions are so few, searching the full input space at
every step is wasteful.  Subspace adversarial training \cite{li2022subspace}
reaches a related conclusion, but extracts its subspace from the optimization
trajectory and works in parameter space; ours is determined by the structure of
the network at the input in question and lies in input space.

A second distinction separates our approach from the certified robustness
literature, where the geometry of a network is controlled by constraining each
layer individually. Parseval networks \cite{cisse2017parseval}, spectral
normalization \cite{miyato2018spectral}, and the orthogonal convolution
constructions of \cite{li2019preventing, trockman2021orthogonalizing,
singla2021skew} all enforce a norm or orthogonality condition on every weight
matrix, so that the Lipschitz constant of the composition is bounded by the
product of the layerwise bounds; the programme is extended to the activations in
\cite{singla2021householder, singla2022improved}. A global guarantee then
follows from purely local conditions, and a certificate is obtained as a
by-product. The cost is that the product bound is loose. For the map
\eqref{eq:affinemap}, the estimate $\|F\| \le \prod_\ell \|W_\ell\|$ holds with
equality only when the dominant singular directions of successive factors align,
which they generically do not, and how far the naive bound sits above the true
constant is documented in \cite{virmaux2018lipschitz, fazlyab2019efficient}. A
network constrained layerwise is thus constrained more tightly than the
robustness objective requires, and the excess is paid in expressivity.

Our approach acts on the product rather than on its factors. The expansion is an
expansion of $F$ itself, so no layer is required to be orthogonal or
norm-bounded, no activation is replaced, and the network is trained in the usual
way. We obtain no certificate, since we impose no condition from which one could
be derived; what we obtain instead is a description of where the composed map is
sensitive that is not conservative by construction. This also underpins the
relation to the prior works using Householder reflections
\cite{singla2021householder}. There the Householder structure is imposed: it is
a design constraint on the activation, satisfied exactly, and the reflection
vectors are trained parameters. Here it is descriptive, read in from a network
trained without any direct restrictions. The theorem of
\cite{singla2021householder} is nonetheless a useful independent justification
for our choice of basis, since it establishes that Householder reflections are
the canonical form for Jacobian transitions across activation boundaries in
piecewise linear networks. Our expansion is thus not an arbitrary change of
coordinates but one aligned with the intrinsic structure of the map being
approximated.

Our experiments on MNIST support the following claims. First, the subspace
identified by the expansion is the one in which adversarial perturbations lie:
projecting perturbations onto its orthogonal complement reduces attack success
from $28.01\%$ to $9.20\%$, whereas removing a random subspace of the same
dimension reduces it only to $25.42\%$ (Section~\ref{sec:inputbasis}).  Second,
training against these directions costs the equivalent of $2.8$ PGD steps per
epoch, an $8.7\times$ reduction relative to $40$-step adversarial training and
less than the cost of 3-step training. Third, over the budget range in which the
expansion is descriptive the resulting models match multi-step adversarial
training: they match the robustness of 3-step PGD training for relative $\ell^2$
budgets $\eps \le 0.02$ and of $40$-step training for $\eps \le 0.012$, with
multi-step training pulling ahead beyond (Section~\ref{sec:results}). The
falling away is consistent with the locality of the expansion, which ceases to
describe the network once the perturbation is large enough to leave the affine
cell.

We regard this parity as the substantive result. Every method in the third group
above obtains its information about the network's local behavior by
differentiating with respect to the input, and it would be natural to assume
that such differentiation is critical to obtaining that information. The
comparison shows otherwise: the geometry is already present in the activation
pattern, and a regularizer that never forms an input derivative reaches similar
performance as one that forms forty. 

We focus our attention on small budgets. The motivation for studying adversarial
examples at all rests on the premise that the perturbation is imperceptible,
although the budgets now conventional in the literature have evolved well beyond
the point where that premise holds. The class we treat is the one the original
problem concerned. We state plainly what is not claimed. We do not report
state-of-the-art robust accuracy at all budgets, and we provide no certificates.
Our experiments are confined to MNIST, and we make no claim that the 
approximation accuracy required to capture the adversarially relevant subspace,
or the budget range over which the expansion remains descriptive, transfers to
higher-resolution data.

\section{Preliminaries}

\subsection{Adversarial examples}
The term \emph{adversarial examples} refers to a broad class of transformations
$T(x)$ of the input data $x$ that cause an incommensurate change in the deep
learning model prediction $f(T(x))$ when compared to $f(x)$.  For classification
tasks, this change refers to the change in the output classification label from
a correct one to an incorrect one.  The literature \cite{tramer2020adaptive}
categorizes adversarial examples into the case when $T(x)$ is small, sometimes
referred to as sensitivity adversarial examples, and the case when $T(x)$ is an
admissible family of group actions, called invariance adversarial examples; e.g.
geometric distortions like rotation or translation. On the other hand,
\emph{natural adversarial examples} \cite{hendrycks2021natural} are
out-of-distribution samples that arise without any well-defined transformation
$T$ and occur in datasets collected without synthetic generation. These examples
are not necessarily minuscule perturbations of the original image, nor described
straightforwardly by geometric invariance.

This work focuses on the specific subclass defined in terms of the $\ell_p$-norm
of $T(x)$, i.e. the sensitivity adversarial examples in the terminology above. 
The precise definition follows, with the slight difference that we measure the
distance in the relative sense.

\begin{defi}[Adversarial examples]
  Given a model $f: \RR^{\ninput} \to \RR^{\noutput}$, and
  an input $x \in \RR^{\ninput}$, the perturbed input $x + \delta x$ is an 
  \emph{adversarial example} if 
  \begin{equation} \label{def:adv}
    \Norm{\delta x}{p} \le \eps \Norm{x}{p}
    \quad \text{ and } \quad
    \softargmax f(x) \ne \softargmax f(x + \delta x).
  \end{equation}
\end{defi}
The notation $\Norm{\cdot}{p}$ denotes the standard $\ell_p$ norm, and we will
focus on $p=2$ throughout this paper, and show some results for $p=\infty$. Note
that in many settings when $p=\infty$ this definition of $\eps$ relative to the
norm of the input reduces to the absolute $\eps$. For example, for grayscale
images in MNIST, $\Norm{x}{\infty} = 1$ or approximately so, reducing the
constraint to an absolute one. Note that we separate out the $\softargmax$
function from the model $f$, which outputs logits.

More generally, adversarial examples for regression tasks are defined by
replacing the misclassification requirement \eqref{def:adv} with
\begin{equation}
\Norm{f(x + \delta x)}{p} \ge \tau \Norm{f(x)}{p},
\end{equation}
for some appropriate percentage threshold $\tau$, which is specially selected
for each domain-specific dataset and learning task. For example, for the
geophysical data set for tsunami waveheight prediction \cite{lrhe2024}, 
a budget of $\eps = 0.005$ relative change in geodetic measurement input causing
$\tau = 0.3$ change in surface elevation was considered adversarial, a
reasonable assumption due to the stability of the underlying physical processes.

We focus on the case where the model $f$ is a feedforward ReLU neural network.
For given $x \in \RR^{\ninput}$ the feedforward neural network model is a
parametrized family of functions 
\begin{equation} \label{eq:feedforward}
  f_\theta(x)
  =
  A_L \, \sigma \, A_{L-1} \, \sigma \, \cdots \, \sigma \, A_1 \, x,
  \quad
  A_\ell z = W_\ell z + b_\ell,
\end{equation}
where the parameters of the network are $\theta = (W_1, ...\,, W_L, b_1, ... \,,
b_L)$, and $\sigma$ here is the Rectified Linear Unit (ReLU) activation 
$\sigma(z) = \max\{0, z\}$ for $z \in \RR$. The left-multiplications by $A_\ell$
and $\sigma$ denote application of an affine map and the entry-wise application
of a ReLU to the hidden state vector, respectively.

Adversarial examples for a trained neural network $f_\theta$ are found by
solving an optimization problem, typically by increasing the misfit loss $\cL:
\RR^{\noutput} \times \{1, ...\,, C\} \to \RR_+$ (e.g. the cross-entropy
function), while keeping the clean input $x$ and trained weights $\theta$ fixed,
that is,
\begin{equation} \label{eq:adv}
  \max_{\Norm{\delta x}{p} \le \eps \Norm{x}{p}} 
    \cL( f_\theta(x + \delta x), y).
\end{equation}
The designated optimization problem and the particular optimization approach is
referred to as a \emph{threat model}. The {\tt AutoAttack} package
\cite{croce2020reliable} implements an accepted benchmark with standard threat
models, such as variants of PGD, the Fast Adaptive Boundary attack
\cite{croce2020minimally}, and the Square attack
\cite{andriushchenko2020square}. To supplement {\tt AutoAttack}, we have
implemented a strong PGD attack with multiple configurations and thousands of
iterations
\cite{goodfellow2015explaining,athalye2018obfuscated,madry2018towards}. We found
that {\tt AutoAttack} reliably outperformed vanilla PGD attacks, so we report
only those results.

\subsection{Low-rank Householder expansion (LRHE)} \label{sec:lrhe}
In a prior work \cite{lrhe2024}, the feedforward neural network $f_\theta$ was
shown to naturally yield an expansion which writes it as a low-rank correction
of an affine mapping. Denote the hidden states as
\begin{equation}
  \left\{
  \begin{aligned}
    z_1 &:= W_1 x + b_1,
    \\
    z_{\ell+1} &:= W_{\ell +1} \sigma z_{\ell} + b_{\ell + 1},
    &&\ell = 1, 2, ...\, , L-1,
  \end{aligned}
  \right.
\end{equation}
then the evaluation of a ReLU $\sigma$ for a given vector $z_\ell \in \RR^\nell$
admits the alternative form
\begin{equation} \label{eq:absrelu}
  \sigma z_\ell = \half ( z_\ell + \abs{z_\ell}).
\end{equation}
The absolute value operation $\abs{\cdot}$ is an orthogonal transformation on
$\RR^{\nell}$ which is a natural candidate to represent as a Householder
reflection, a rank-one perturbation of the identity of the form
\begin{equation}
  \abs{z_\ell}
  =
  (I - 2 v_\ell v_\ell^\top) z_\ell,
\end{equation}
where the vector $v_\ell$ is determined modulo sign by the two conditions
\begin{equation}
  v_\ell \parallel z_\ell - \abs{z_\ell},
  \qquad
  \Norm{v_\ell}{2} = 1.
\end{equation}
Inserting this formulation into \eqref{eq:absrelu} and into the neural network
\eqref{eq:feedforward}, we obtain the product
\begin{equation}
  f_\theta (x)
  =
  A_L (I - v_{L-1} v_{L-1}^\top) A_{L-1} 
    \cdots A_2  (I - v_1 v_1^\top) A_1 x.
\end{equation}
Expanding the product on the RHS, one obtains $2^{L-1}$ terms. All but the
input-independent term $F_0$ below are input-dependent and rank one. That is, writing
\begin{equation}
  f_\theta(x)
  =
  F_0 x + F_\sigma x,
\end{equation}
the affine maps $F_0$ and $F_\sigma$ are given by
\begin{equation} \label{eq:expand}
  F_0 = A_L A_{L-1} \cdots A_2 A_1,
  \Aand
  F_\sigma = \sum_{\beta = 1}^{2^{L-1} - 1} F_\beta,
\end{equation}
and every term in the sum $F_\sigma$ over the index $\beta$ is of the form
\begin{equation}
  A_L \cdots A_{\ell+1} v_\ell v_\ell^\top A_\ell \cdots A_1,
\end{equation}
where there is at least one rank-one term $v_\ell v_\ell^\top$ present as a
factor, so the product as a whole is rank-one by the rank inequality for
products. Furthermore, the column and row spaces of $F_\sigma$ are each spanned
by $L-1$ vectors,
\begin{equation}
  W_L \cdots W_{\ell+1} v_\ell,
  \Aand
   W_1^\top \cdots W_\ell^\top v_\ell,
  \Ffor
  \ell = 1, \cdots, L-1.
\end{equation}
Here the terms \emph{column} and \emph{row space} of the affine map $F_\sigma$ refer to the
column and the row space of the linear part of $F_\sigma$, respectively;
similarly the rank of $F_\sigma$ refers to the dimension of the column or row
space of $F_\sigma$.

Hence the crucial property of this expansion is that the affine transformation
$F_\sigma$ has rank at most $L-1$, and is therefore low-rank whenever $L-1$ is
small relative to the layer widths. This expansion is called the low-rank Householder expansion (see
Theorem 2.4 in \cite{lrhe2024}).

\begin{defi} \label{def:lrhe}
The \emph{Low-Rank Householder Expansion (LRHE)} of the neural network
$f_\theta$ is the expansion given by the substitution of the activation $\sigma$
after the $\ell$-th layer by the rank-one perturbation of the identity $I -
v_\ell v_\ell^\top$ which reads
\begin{equation}
  f_\theta(x) =  F_0(x) + F_\sigma(x),
\end{equation}
where $F_0$ is an input-independent affine transformation and $F_\sigma$ is an
input-dependent affine transformation and satisfies $\text{rank}(F_\sigma )\le
L-1$.  The column and row spaces of low-rank affine transformation $F_\sigma$
are
\begin{equation} \label{eq:rowcol}
  \begin{aligned}
  \col(F_\sigma)
  &=
  \Span\{
    W_L W_{L-1} \cdots  W_{\ell+2} W_{\ell+1} v_\ell
    \mid
    \ell = 1, 2, ...\,, L-1
  \},
  \\
  \row (F_\sigma) 
  &=
  \Span\{
    W_1^\top W_2^\top \cdots W_{\ell-1}^\top W_\ell^\top v_\ell
    \mid
    \ell = 1, 2, ...\,, L-1
  \}.
  \end{aligned}
\end{equation}
\end{defi}

For a fixed input $x$, the representation of a ReLU using a Householder
reflection in LRHE reveals a new low-rank structure in $F_\sigma$ that pertains
purely to the nonlinear behavior of the neural network, independently from the
linear analysis of $F_0$. Note that this representation of the ReLU is distinct
from so-called \emph{Householder activations} \cite{singla2021householder},
whose definition involves Householder reflections.

The core idea motivating this work is the hypothesis that LRHE and adversarial
examples are closely related; LRHE provides a way to explain how the adversarial
examples can arise \cite{lrhe2024}. In LRHE the linear part of the
input-dependent low-rank term $F_\sigma$ can be factorized into a singular value
decomposition of the linear part,
\begin{equation} \label{eq:Fssvd}
  F_\sigma z = b_\sigma 
  +
  \sum_{i=1}^{L-1} d_i \zeta_i \xi_i^\top z,
  \quad
  b_\sigma = F_\sigma 0,
  \,\,
  d_i \in \RR_+,
  \,\,
  \zeta_i \in \RR^{\noutput},
  \,\,
  \xi_i \in \RR^{\ninput}.
\end{equation}
Instability can arise when $d_1 \gg 1$ even if $F_0$ is well-conditioned, so
that $\Norm{F_0 \, \delta x}{p} \sim \Norm{\delta x}{p}$. When the perturbation $\delta x$ is in the direction of $\xi_1$,
it is possible to have
\begin{equation}
\Norm{\delta x}{p} \ll 1
\Aand
\Norm{F_\sigma \; \delta x}{p} \gg 1.
\end{equation}
This decouples the linear conditioning of $F_0$ from the nonlinear conditioning
of $F_\sigma$ and of the neural network $f_\theta$.

There are previous approaches aimed at making $F_0$ well-conditioned via the
submultiplicative inequality $\Norm{F_0}{p} \le \prod_\ell \Norm{W_\ell}{p}
\lesssim 1$ considered already in \cite{szegedy2014intriguing}, for example by
imposing weight decay during training. However, this reduces to penalizing
$\Norm{W_\ell}{}$, i.e. weight decay, and standard regularization does not
generally lead to adversarial robustness; more sophisticated considerations are
necessary, e.g. \cite{singla2021householder,singla2022improved} as discussed in
the introduction.

\section{Adversarial training based on LRHE} \label{sec:advtrain}

\subsection{The min-max formulation}
\emph{Adversarial training} \cite{goodfellow2015explaining,madry2018towards}
refers to training strategies that broaden the scope of the usual empirical risk
over the data distribution $\cD$
\begin{equation}
  \min_{\theta} \EE_{(x, y) \sim \cD}
    \left[ \cL( f_\theta(x), y) \right] 
\end{equation}
to cover worst-case or near-worst-case adversarial examples. For example, one
takes the minimization problem to be
\begin{equation} \label{eq:advtrain}
  \min_{\theta}
  \EE_{(x, y) \sim \cD}
    \left[
      \max_{\Norm{\delta x }{p} \le \eps \Norm{x}{p}}  \cL( f_\theta(x + \delta x), y)
    \right].
\end{equation}
Inner maximization has no explicit solution and is commonly approximated by $K$
steps of projected gradient ascent. Each step requires a gradient of the loss
with respect to the input, and the entire cost of adversarial training in excess
of ordinary training is the cost of this search. What the search returns,
however, is a direction, and in the regime where the network is affine that
direction is not arbitrary: the maximizer of the linearized inner objective over
$\Norm{\delta x}{p} \le \eps \Norm{x}{p}$ is $\eps$ times the unit vector most
amplified by the model in the metric induced by the output-space loss gradient.
PGD is thus an iterative procedure for locating a direction that is already
determined by the local action of the network.

The LRHE supplies such directions directly. Its right factors $\xi_1, \dots,
\xi_{L-1}$ in the SVD of $F_\sigma$ \eqref{eq:Fssvd} are, by construction, the
directions in which the composed map amplifies most strongly, and they are
obtained from the activation pattern without any differentiation with respect to
$x$. 

We first motivate our training objective conceptually. The key ingredient is the
restriction of the inner maximization of \eqref{eq:advtrain} by a penalty
evaluated on this fixed set of directions by the orthonormal vectors
$(\xi_\ell)_{\ell}$,
\begin{equation}
  \min_{\theta}\;
  \mathbb{E}_{(x,y)}\Bigl[
    \mathcal{L}\bigl(f_\theta(x), y\bigr)
    + \lambda \sum_{\ell=1}^{L-1}
      \mathcal{L} (f_\theta(x + \eps \xi_\ell), \softargmax f_\theta(x) )
  \Bigr],
  \label{eq:lrhe-objective}
\end{equation}
so that the min-max problem collapses to a minimization. The structure of
adversarial training is retained, that is, the model is penalized for its
response to perturbations of size $\eps$ in the directions that most affect it.
In contrast, the search is eliminated, and with it the $K$ input-gradient
computations that dominate its cost.

Following these discussions, we propose a new training procedure based on the
LRHE introduced above. A na\"ive approach would penalize the largest singular
value $d_1$ of $F_\sigma$ \eqref{eq:Fssvd}, however, the direct computation of
the SVD of $F_\sigma$ would involve accumulating $2^{L-1}$ terms in the
expansion \eqref{eq:expand}, which is prohibitive. Instead, we penalize a basis
for the column space of $F_\sigma$ \eqref{eq:rowcol}. 

We keep the bias terms in the affine maps, assuming that the large singular
values in the linear part of $F_\sigma$ \eqref{eq:Fssvd} make the linear part
dominant. The \emph{output basis vectors} are computed during a forward pass
\begin{equation}
  \phi_\ell (x)
  :=
  A_L A_{L-1} \cdots A_{\ell+1} 
  v_\ell v_\ell^\top  
  A_{\ell-1} \sigma \cdots \sigma A_1 x.
  \qquad
  \ell = 1, ...\,, L-1.
\end{equation}
The vector $\phi_\ell$ has the alternative expression,
\begin{equation} \label{eq:output_basis}
  \phi_\ell (x)
  =
  A_L A_{L-1} \cdots A_{\ell+1} (\sigma z_\ell - z_\ell),
  \quad
  \ell = 1, ...\,, L-1.
\end{equation}
The input basis, i.e. the basis for the row space in \eqref{eq:rowcol}, is
computed by reversing the computation for the output basis $\phi_\ell$
\eqref{eq:output_basis},
\begin{equation} \label{eq:psi}
  \psi_\ell(x) 
  =
  W_1^\top W_2^\top \cdots W_{\ell - 1}^\top W_\ell^\top 
  (\sigma z_\ell - z_\ell),
  \quad
  \ell = 1, ...\,, L-1.
\end{equation}
This computation gives us the row and column basis of $F_\sigma$ modulo the bias
terms; the output basis is computed with the affine maps and therefore lies in 
$\col(F_\sigma)$ only modulo the bias offset, whereas the input basis is exact.
Neither basis coincides with the singular vectors of $F_\sigma$, however. Note the
discussion on their relation in \cite{lrhe2024} which shows the singular vectors
can be estimated empirically by a sparse sampling of the terms in the sum
\eqref{eq:expand}. 

Finally, we approximate the response of perturbing the input by $\eps \xi_i$ in
\eqref{eq:lrhe-objective} via the approximation
\begin{equation}
  f_\theta(x + \eps \xi_\ell) - f_\theta(x)
  \approx
  \eps (F_0 + F_\sigma ) \xi_\ell,
\end{equation}
via the nonlinear part $F_\sigma$ of LRHE, and based on our discussion regarding
how the adversarial examples can arise from the viewpoint regarding
\eqref{eq:Fssvd}, our priority is to suppress the dominant term 
\begin{equation}
  d_\ell \zeta_\ell + b_\sigma = F_\sigma \xi_\ell \gg F_0 \xi_\ell,
\end{equation}
that is, the singular values $d_\ell$ and the corresponding left singular
vectors $\zeta_\ell$. As already mentioned, these are hard to compute directly,
hence we compute $(\phi_\ell)_\ell$ instead. From this alternative, we
want to penalize the change
\begin{equation}
 \Norm{f_\theta(x + \eps \xi_\ell) - f_\theta(x)}{p} \lesssim \eps d_\ell \Norm{\zeta_\ell }{p}
\end{equation}
indirectly by penalizing the size of $\phi_\ell$. For classification tasks, it
is natural to penalize the deviation $f_\theta(x) + \gamma \phi_\ell$ so long as
the label does not change, leading to the penalization 
\begin{equation}
  \cL(f_\theta(x) + \gamma \phi_\ell, \softargmax f_\theta(x)),
\end{equation}
with some scaling $\gamma$.

Putting it all together, we propose our adversarial training as the regularized
problem
\begin{equation} \label{eq:obj}
    \min_{\theta} 
      \EE_{(x, y) \sim \cD}
        \left[ \cL( f_\theta(x), y) 
      +
      \lambda
      \cR(\theta, x; \gamma) \right],
\end{equation}
where the regularization term penalizes the change in the final prediction,
\begin{equation} \label{eq:reg}
  \cR(\theta, x; \gamma)
  =
  \sum_{\ell=1}^{L-1} \cL(f_\theta(x) + \gamma \phi_\ell, \softargmax f_\theta(x)),
  \quad
  \gamma \in (-1, 1),
\end{equation}
where $\lambda$ and $\gamma$ are hyper-parameters.

The computation of these bases $(\phi_\ell)_{\ell}$ and $(\psi_\ell)_{\ell}$ is
straightforward; one computes the difference of the pre- and
post-activation-states, then applies all the subsequent affine layers to obtain
the output basis, or all the preceding transposed weights for the input basis. 

\begin{figure}
  \includegraphics[width=1.0\textwidth]{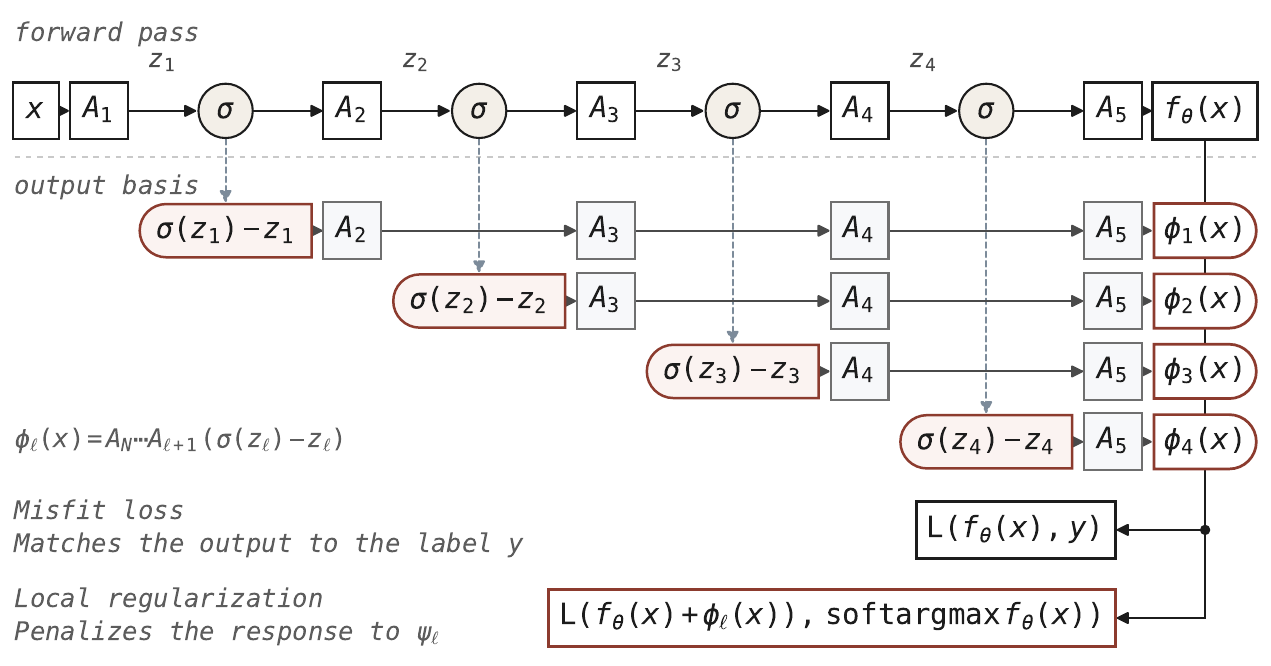}
  \caption{Diagram of a forward pass that computes the output basis 
  $(\phi_\ell)_\ell$ \eqref{eq:output_basis}.
  The computation of the output basis
  involves additional partial forward passes that omit the nonlinearities.  The
  output basis is then used in the regularization term $\cR$ \eqref{eq:reg} to
  penalize large changes in the output logits in the direction of the output
  basis.}
  \label{fig:basisforward}
\end{figure}

During training of the regularized problem \eqref{eq:obj}, $f_\theta$ is
evaluated as usual until the $\ell$-th layer, then only the affine mappings are
applied in the subsequent layers omitting the nonlinearities. This propagates
the nonlinear correction $\sigma z_{\ell+1} - z_{\ell+1}$ through the subsequent
layers. At each gradient computation, only one forward and one backward pass are
required. The cost of forward evaluation of this regularizer is $\cO(L)$ times
the standard evaluation of the forward model $f_\theta$, and one backward pass
is required to compute the gradient of the loss function with respect to
$\theta$ during training. For models with $L$ not too large, however, minimizing
the objective \eqref{eq:obj} is more efficient than the standard adversarial
training \eqref{eq:advtrain} which would require multiple forward-backward
passes with respect to both parameters $\theta$ and input $x$ within the
inner-max-loop (PGD iterations).

Figure~\ref{fig:basisforward} depicts a diagram showing the flow of computation
for the output basis $(\phi_\ell)_\ell$ and the regularization term.

\subsection{Remarks on the training procedure}
\begin{figure}
  \centering
  \includegraphics[width=0.65\textwidth]{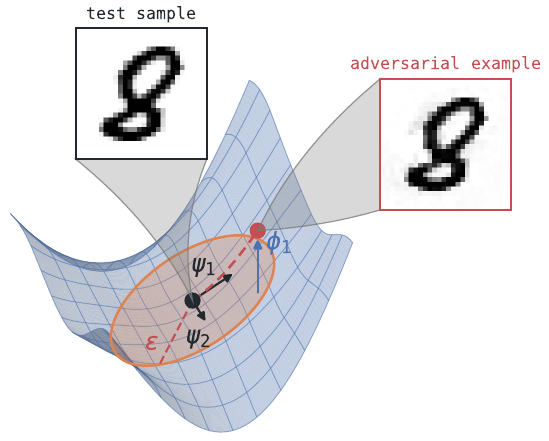}
  \caption{An illustration relating LRHE input basis to adversarial examples.
  The surface denotes the model logit output for the incorrect class, which
  increases quickly in one of the input-basis directions. In the linear regime,
  when the input is perturbed in the direction of $\psi_1$ the logit changes in
  the direction of $\phi_1$ resulting in a misclassification.}
  \label{fig:manifold}
\end{figure}

Let us discuss the design of this regularizer. In the LRHE in
Definition~\ref{def:lrhe}, the input and output bases are coupled via the
input-dependent affine map $F_\sigma$. If these bases are indeed closely related
to adversarial examples as surmised in Section~\ref{sec:lrhe}, then the
adversarial examples occur because a perturbation $\psi \in
\Span (\psi_\ell)_\ell$ causes the output $f_\theta (x + \psi)$ to have a
significant component in $\Span \{ \phi_\ell \}_\ell$. In other words,
\begin{equation}
  \phi_* = \argmin_{ \phi \in \Span(\phi_\ell )_\ell}
   \Norm{f_\theta(x + \psi) - f_\theta(x) - \phi}{p}
\end{equation}
would form the main component of the output perturbation that causes
misclassification. Figure~\ref{fig:manifold} shows an illustration of the input
and output basis and their roles in increasing the logit of an incorrect class.

Note that the computation of $\phi_\ell$ involves not just the weights of a
single layer, but their compositions. The regularization controls the stability
of the composed mappings
\begin{equation}
  A_L A_{L-1} \cdots A_{\ell+1}
  \quad
  \Ffor \ell = 1, 2, \dots, L-1,
\end{equation}
rather than the behavior of the individual $A_\ell$, or the usual per-layer
weight norm $\Norm{W_\ell}{}$.

Additional differences of the proposed approach \eqref{eq:obj} from the standard
adversarial training \eqref{eq:advtrain} should be acknowledged. The LRHE
directions depend on the input and the network but not on the label: they
describe where the map is sensitive, not where the decision boundary is nearest.
The perturbations we penalize are therefore not adversarial in the strict sense
of maximizing the loss, and the correspondence between the two objectives is
exact only to the extent that loss-increasing directions coincide with
amplifying ones. Moreover, projected gradient ascent is free to follow the loss
surface across activation boundaries, whereas the expansion describes a single
affine cell close to the input. 

The mechanics of the training proposed above are very general and flexible; the
formulation \eqref{eq:obj} is merely the most straightforward of numerous
possibilities. As explained already, the
computation of the output basis used in the regularization term \eqref{eq:reg}
is made up of two simple steps: (1) compute the difference between the states
before and after the activation, then (2) apply all the subsequent layers while
omitting the activation. To compute the input basis, simply apply the
transposes of all the preceding layers while ignoring the activations. These
operations generalize to other architectures that are not feedforward, for
example, to architectures with attention layers \cite{vaswani2017attention}.
While this manuscript focuses on adversarial training that exploits the output
basis only, similar training strategies can be derived using the input basis.
Another important consideration is that, if small-norm adversarial examples
are not present, the input basis can represent the directions along which the
output varies most sensitively.  Due to the ease
with which the input basis can be computed, the basis can be useful for
sensitivity analysis of deep learning models.

\section{Convolutional Neural Network for MNIST}

We perform experiments using the MNIST data set. This section describes the
model and the attack used throughout, and then treats a preliminary that the
expansion requires: max-pooling layers must be rewritten before the LRHE can be
computed for a convolutional network.

\begin{figure}
  \centering
  \includegraphics[width=0.92\textwidth]{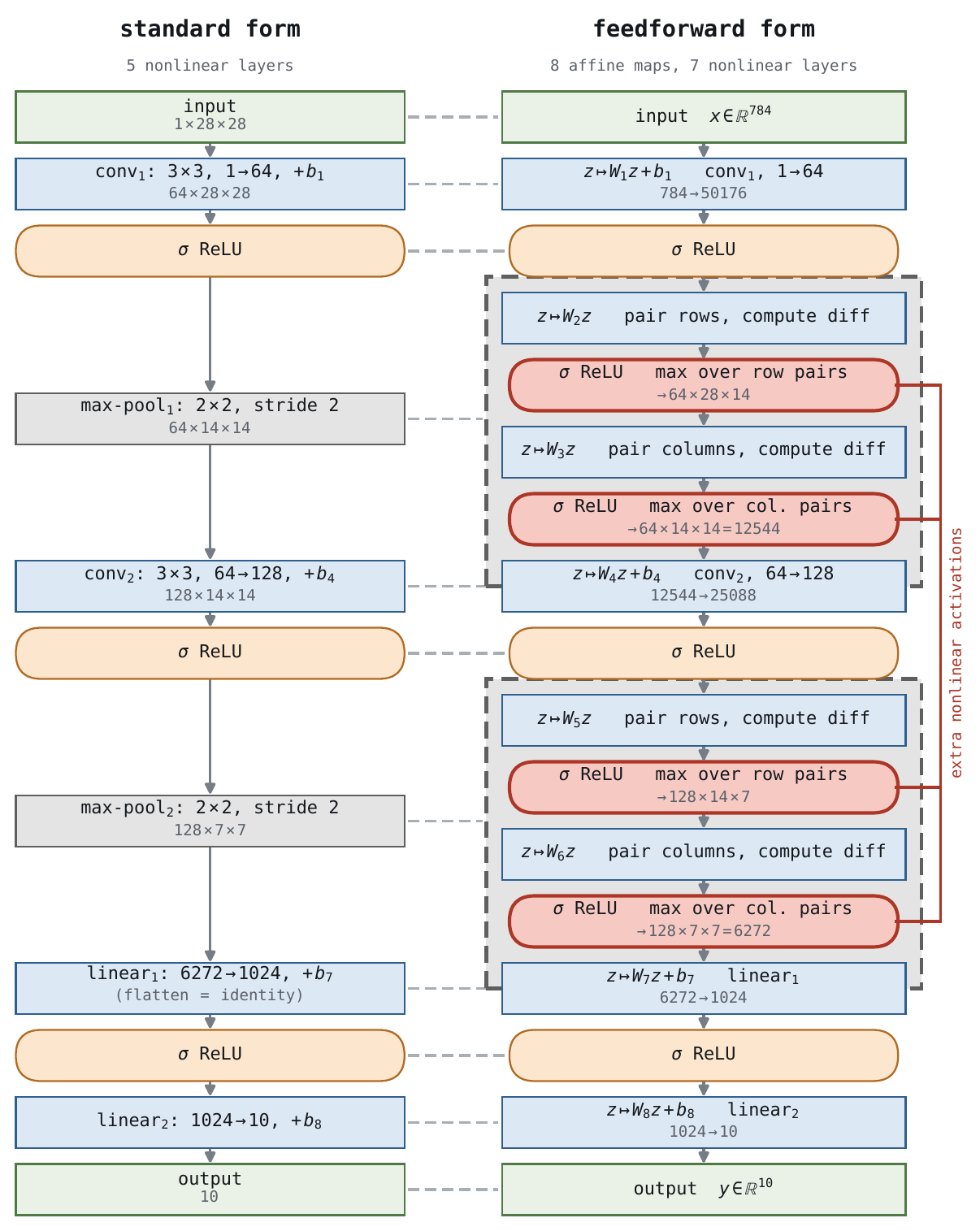}
  \caption{A diagram of the CNN architecture used for the MNIST
  experiments (left). The feedforward form of the same architecture, where the
  max-pooling layers are converted into a ReLU network, is shown alongside
  (right).  Four new ReLUs are added to the feedforward form, which implies
  additional output basis vectors $\phi_\ell$ will be computed.}
  \label{fig:arch}
\end{figure}

\subsection{Clean training and adversarial examples}

To prepare our experiments, we train a convolutional neural network (CNN)
using standard training. The architecture is two convolutional layers, each
followed by ReLU and two-dimensional max-pooling, and then two linear layers.
The hyperparameters for each layer are
\begin{itemize}
  \item Convolutional layer 1: 1 input channel, 64 output channels
  \item 2D max-pooling layer 1: stride 2, kernel size 2
  \item Convolutional layer 2: 64 input channels, 128 output channels
  \item 2D max-pooling layer 2: stride 2, kernel size 2
  \item Linear layer 1: input dim. $128 \cdot 7 \cdot 7 = 6272$, output
    dim. 1024
  \item Linear layer 2: input dim. 1024, output dim. 10.
\end{itemize}
A diagram depicting this architecture is shown in Figure~\ref{fig:arch}. We use
the cross entropy loss and train with the Adam optimizer \cite{Kingma2015AdamAM}
at learning rate {\tt 1e-3} for 20 epochs. The trained model achieves $99.48\%$
test accuracy.

We conduct attacks on this model, at the relative norm sizes of
Definition~\ref{def:adv}, with {\tt AutoAttack} \cite{croce2020reliable}. All
hyperparameters of the ensemble are fixed, so no tuning is performed on our
behalf, and the model is deterministic at inference, so the standard version of
the ensemble is the appropriate one. The ensemble attack raised none of its
automatic warnings regarding the reliability of the evaluation. For example, at
a relative $\ell^2$ radius of $\eps=0.075$, the ensemble reduced the accuracy of
the clean model to $88.40\%$.  We report the attack result for the clean model
below in Section~\ref{sec:results}, in Table~\ref{tab:robust}.

To confirm that the attack above is not merely a weak one, we additionally
attack the clean model with $20$ runs of $1{,}000$ PGD iterations.  These
parameters differ somewhat from those used commonly in the $\ell^\infty$ case;
our hyperparameter exploration found that large PGD step sizes yielded more
examples. The attack had a success rate of $9.79\%$ at a relative $\ell^2$
radius of $\eps=0.1$. This is significantly weaker than the $28.01\%$ success
rate of {\tt AutoAttack}, but comparable to the $11.08\%$ that {\tt AutoAttack}
achieves at the smaller budget $\eps=0.075$.

\begin{figure}
  \includegraphics[width=0.65\textwidth]
    {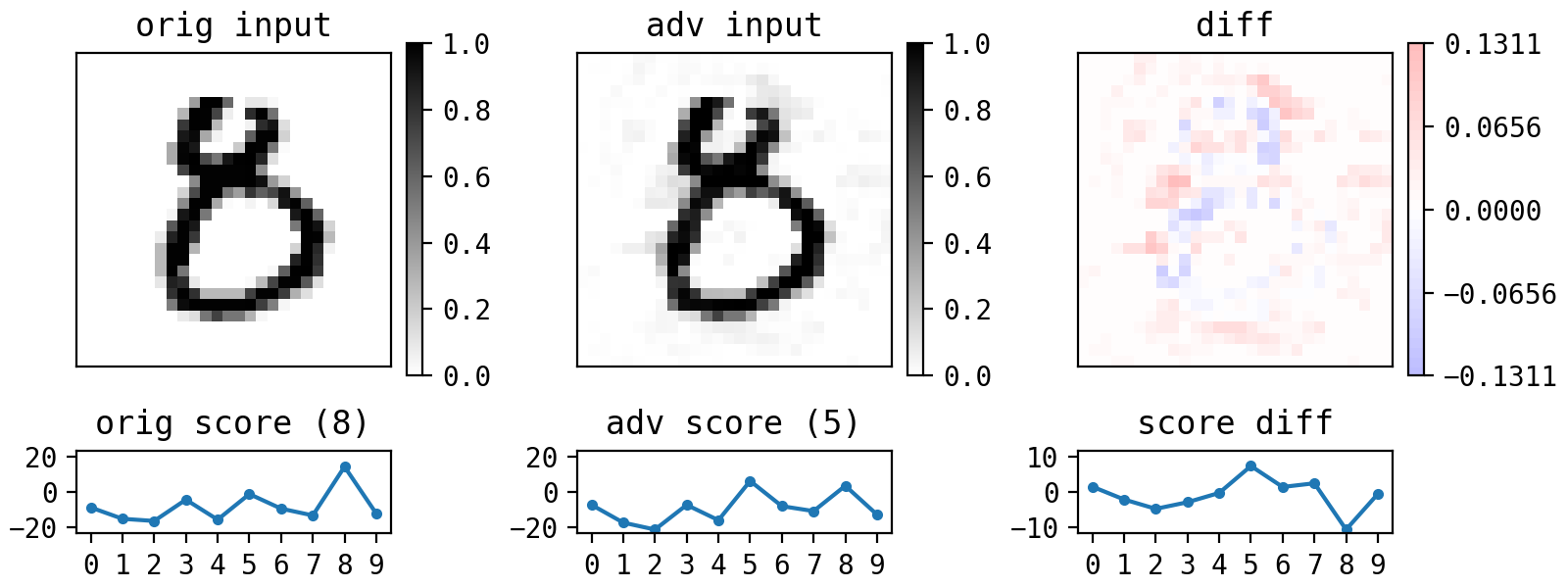}
  \includegraphics[width=0.65\textwidth]
    {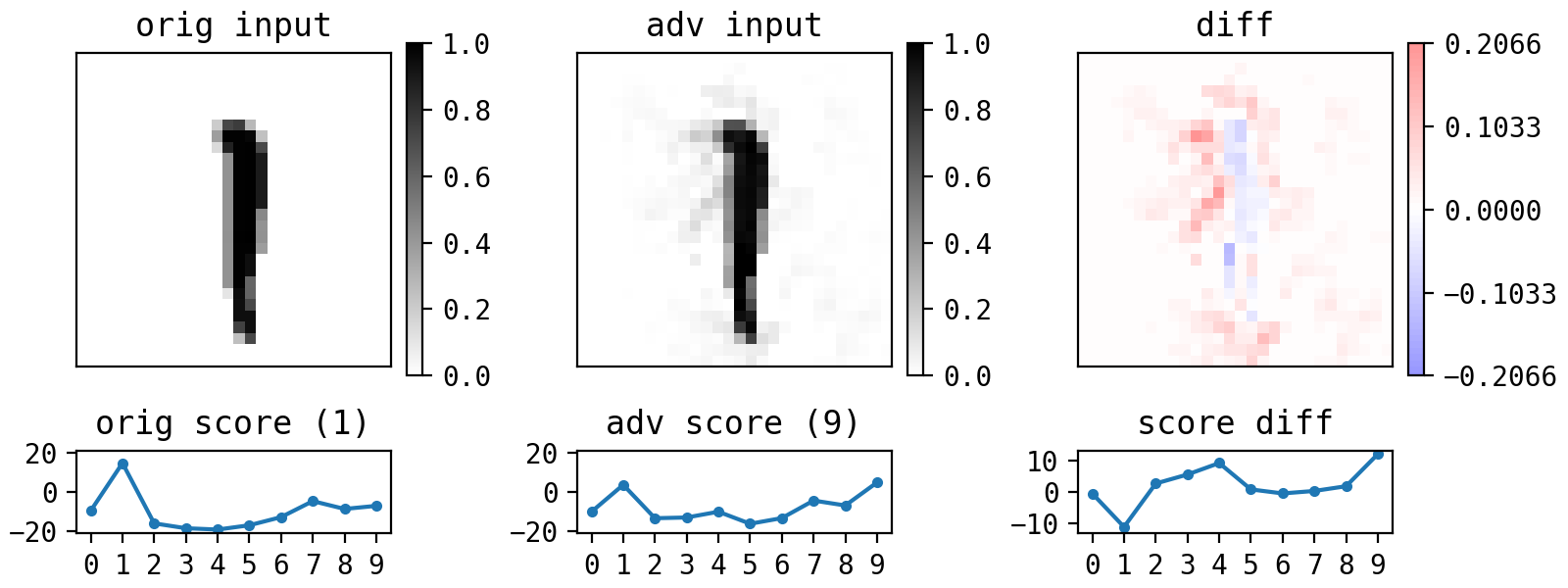}
  \includegraphics[width=0.65\textwidth]
    {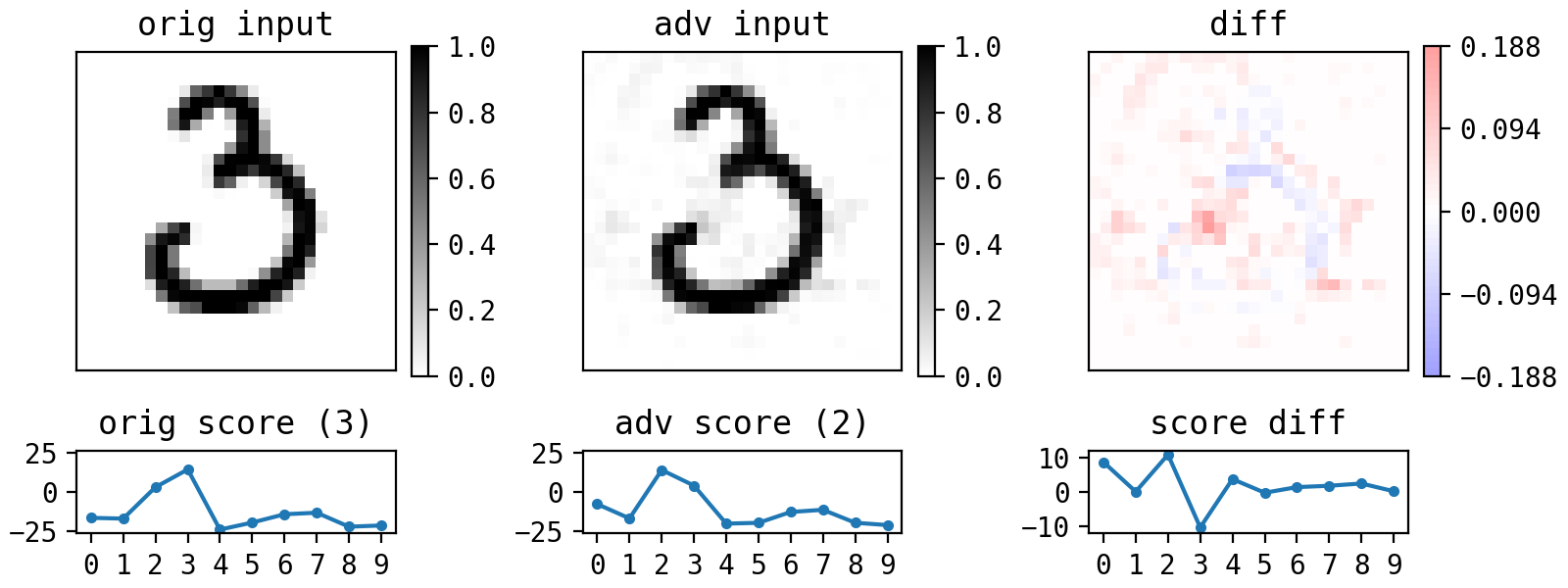}
  \includegraphics[width=0.65\textwidth]
    {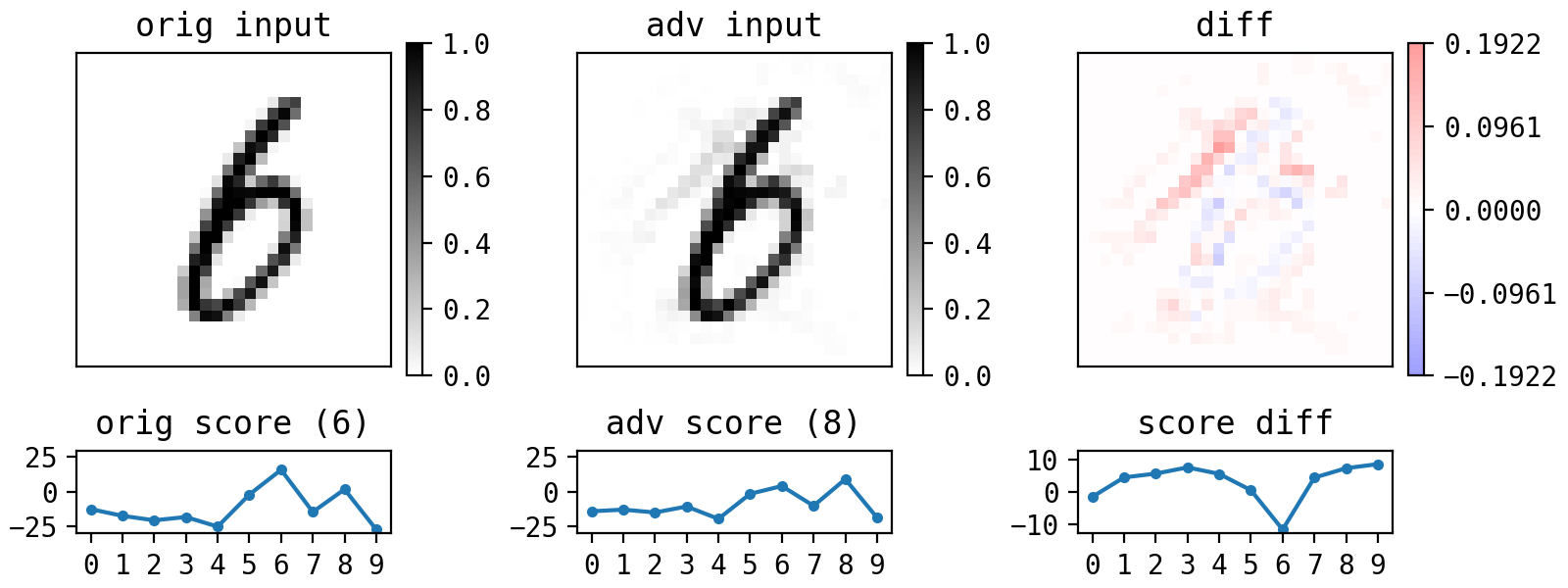}
  \includegraphics[width=0.65\textwidth]
    {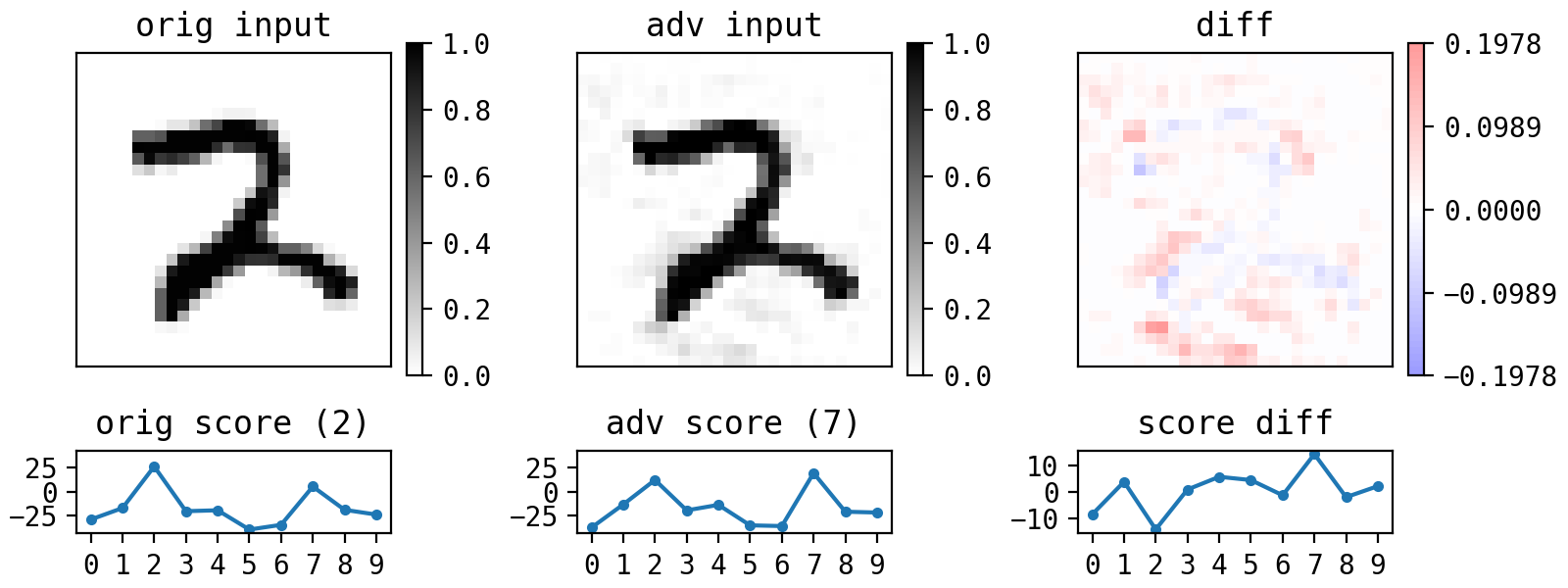}
  \caption{Adversarial examples found for the standard CNN model using PGD
  with relative $\ell^2$ norm $\eps = 0.1$.}
  \label{fig:orig_adv}
\end{figure}

Some of the adversarial examples found are shown in Figure~\ref{fig:orig_adv}.
They are considerably smoother than the $\ell^\infty$-norm adversarial examples
obtained at larger $\eps$ \cite{goodfellow2015explaining}, and appear as though
the original hand-written digits had been written in pencil and then smudged, a
plausible physical scenario that can occur in real data.

Computing the LRHE for this model is not immediate, because the expansion
presumes a particular form of the network. We require the model in feedforward
form, by which we mean a strict alternation of affine maps and a single
pointwise nonlinearity \eqref{eq:feedforward}, where each $A_\ell$ is affine and
$\sigma$ acts entrywise. The expansion of Section~\ref{sec:lrhe} is built
directly on this structure: the correction $\sigma(z_\ell) - z_\ell$ is defined
relative to the pre-activation $z_\ell$ of each nonlinear layer, and the basis
functions $\psi_\ell$ are obtained by propagating that correction through the
remaining affine maps. A layer that is neither affine nor a pointwise
application of $\sigma$ therefore has no place in the expansion until it is
rewritten.

For most of the architecture above this requirement is already met.
Fully-connected and convolutional layers are affine in the input and are
absorbed into a single $A_\ell$; strides, padding, and channel mixing change the
matrix but not its affineness. Average pooling is likewise affine and would need
no special treatment. Max-pooling is the only component of the standard CNN that
is not affine, and it is the subject of the remainder of this section.

\subsection{Max-pooling layers in feedforward form}
\label{sec:maxpool-feedforward}

There are several admissible ways of rewriting a max-pooling layer in
feedforward form. We adopt the one that expresses the pairwise maximum as a
single nonlinear layer, and then build the pooling operator by composition.
Both of the representations below rest on elementary properties of $\sigma$.
The first recovers a linear passthrough from a nonlinearity,
\begin{equation}
  z \;=\; \sigma(z) - \sigma(-z)
  \qquad \text{for all } z \in \mathbb{R},
  \label{eq:relu-identity}
\end{equation}
and is what allows a coordinate to survive a nonlinear layer unchanged; it costs
two rows rather than one, since the network form \eqref{eq:feedforward} admits
no skip connections. The second expresses the maximum as a shifted rectifier,
\begin{equation}
  \max(a, b) = a + \sigma(b - a).
  \label{eq:max-identity}
\end{equation}
Combining \eqref{eq:relu-identity} and \eqref{eq:max-identity} with $a = x_1$
and $b = x_2$ gives the representation
\begin{equation}
  \max(x_1, x_2)
  =
  \begin{bmatrix}
      1 & -1 & 1
  \end{bmatrix}
  \sigma
  \left(
    \begin{bmatrix}
      1 & 0
      \\
    -1 & 0
      \\
    -1 & 1
    \end{bmatrix}
    \begin{bmatrix}
      x_1 \\ x_2
    \end{bmatrix}
  \right),
  \label{eq:pairwise-max}
\end{equation}
which is exact for all $x_1, x_2 \in \mathbb{R}$. Reading the three rows of the
inner matrix in turn, the pre-activations are $z = (x_1,\, -x_1,\, x_2 - x_1)$,
so the outer row vector forms
\begin{equation}
  \sigma(x_1) - \sigma(-x_1) + \sigma(x_2 - x_1)
  \;=\; x_1 + \sigma(x_2 - x_1)
  \;=\; \max(x_1, x_2),
\end{equation}
where the first two terms collapse by \eqref{eq:relu-identity} and the remainder
is \eqref{eq:max-identity}. The first two rows carry $x_1$ past the
nonlinearity; only the third performs the comparison.

This representation is not unique. For example, note that in a CNN the input to
a pooling layer is the output of a ReLU, hence nonnegative, and the passthrough
may then be had for a single row rather than two. For $x_1, x_2 \ge 0$ we have
$x_1 = \sigma(x_1)$, and \eqref{eq:max-identity} gives the two-row
representation
\begin{equation}
  \max(x_1, x_2)
  =
  \begin{bmatrix}
      1 & 1
  \end{bmatrix}
  \sigma
  \left(
    \begin{bmatrix}
      1 & 0
      \\
    -1 & 1
    \end{bmatrix}
    \begin{bmatrix}
      x_1 \\ x_2
    \end{bmatrix}
  \right),
  \label{eq:pairwise-max-nonneg}
\end{equation}
with pre-activations $z = (x_1,\,x_2 - x_1)$. This is a more economical form,
but we use \eqref{eq:pairwise-max} in practice without assuming the pooling
layers' input is known to be nonnegative. One may also choose a symmetric
version of \eqref{eq:max-identity}; we do not explore these options here.

The pooling operator itself is assembled from these pairwise maxima. Let $P$
denote the patch-extraction operator that gathers, for each output position, the
$s$ inputs lying in its pooling window. For non-overlapping windows $P$ is a
permutation followed by a reshape; for stride smaller than the window it
duplicates entries. In either case $P$ is linear, and is absorbed into the
adjacent affine map at no cost. The pooling layer is then the $s$-way maximum
applied independently to each extracted window. Such a maximum is obtained by
composing pairwise maxima in a balanced binary tree, requiring $s - 1$
applications of \eqref{eq:pairwise-max} arranged in $\lceil \log_2 s \rceil$
successive nonlinear layers. All windows of a given feature map are processed in
parallel, so the maxima at a common depth of the tree occupy a single layer of
\eqref{eq:feedforward} whose width is the sum over windows. For the usual $s =
2$ the tree has depth one, and pooling along a single axis costs exactly one
nonlinear layer.

Two-dimensional pooling reduces to this case, since the maximum over an $s
\times t$ window is separable, i.e.
$\max_{1 \le i \le s,\, 1 \le j \le t} x_{ij}
=
\max_{1 \le i \le s} ( \max_{1 \le j \le t} x_{ij} ),$
so the layer is realised by applying the one-dimensional construction along each
axis in turn, with a permutation between the two stages that is again absorbed
into the intervening affine map. For the standard $2 \times 2$ window this gives
two nonlinear layers, one per axis, and hence three affine maps $A_\ell$ in the
alternating form \eqref{eq:feedforward}. In the assembled network the outer two
of these merge with the affine maps of the preceding and following layers, so
the net effect of a $2 \times 2$ max-pool is to insert two applications of
$\sigma$. Equivalently, one may treat the $st$ window entries as a single flat
tree of depth $\lceil \log_2 st \rceil$, which for the window sizes in common
use coincides with the separable count and is never larger.

Figure~\ref{fig:arch} shows a diagram of the feedforward form.

The width of the resulting layers can be counted. Let the input to the pooling
layer have $C$ channels and spatial extent $H \times W$, with a non-overlapping
$2 \times 2$ window. Using \eqref{eq:pairwise-max}, the first stage produces
$CH(W/2)$ maxima at three rows each, for a pre-activation width of
$\frac{3}{2}CHW$; the second stage produces $C(H/2)(W/2)$ maxima, for a width of
$\frac{3}{4}CHW$.  The pooling layer in the general form \eqref{eq:pairwise-max}
would cost $\tfrac{9}{4} CHW$ pre-activations in total, against $CHW$ entering
it.  Using \eqref{eq:pairwise-max-nonneg} would reduce the total to
$\tfrac{3}{2} CHW$, a third less.

Both \eqref{eq:pairwise-max} and \eqref{eq:pairwise-max-nonneg} are identities,
not approximations. The rewritten network computes precisely the same function
as the original CNN, and no retraining is required. Only the
representation changes, and with it the number of nonlinear layers used in the
expansion. Each nonlinear layer introduced by a pooling operator carries
its own correction and therefore contributes an additional basis function
$\psi_\ell$; a network with two $2 \times 2$ pooling layers gains four
basis functions, which is counted towards $L$ of Section~\ref{sec:lrhe}. 

\section{LRHE basis results}\label{sec:inputbasis}

\subsection{LRHE basis attack} 
We explore the relation between the LRHE input basis $(\psi_\ell)_\ell$ and
adversarial examples. 

First, we test if the input perturbations using the basis functions themselves
directly lead to adversarial examples, by perturbing the clean model
\begin{equation} \label{eq:lrhe-attack}
  f_\theta \left(x \pm \alpha \cdot \eps \Norm{x}{p} \cdot \frac{\psi_\ell}{\Norm{\psi_\ell}{p}} \right),
  \quad
  \ell = 1, 2, ...\,, L-1, 
\end{equation}
for step-sizes $\alpha = 0.5, 1, 1.5, ...\,,3.0$, with $\eps=0.1$ and checking
whether the predicted label switches to an incorrect one. We do not clip the
resulting perturbed input to the range $[0, 1]$ since that would alter the
linear scaling with respect to $\alpha$, so the final values veer outside of the
unit interval by a small amount.

\begin{table}
  \begin{tabular}{r|r}
    \hline
    $\alpha$ & Adv. acc.    \\
    \hline
    $0.5$  &   99.04 (0.44) \\ 
    $1.0$  &   98.84 (0.64) \\ 
    $1.5$  &   98.70 (0.78) \\ 
    $2.0$  &   98.16 (1.32) \\ 
    $2.5$  &   97.90 (1.58) \\ 
    $3.0$  &   96.45 (3.03) \\ 
    \hline
    Total  &   91.69 (7.79)
    \\
    \hline
  \end{tabular}
  \caption{Adversarial accuracy of the clean model after LRHE basis attack 
  for each step-size $\alpha$ \eqref{eq:lrhe-attack}. Difference from clean
  accuracy of 99.48\% is in the parentheses.}
  \label{tab:lrhe-attack}
\end{table}

\begin{figure}
  \includegraphics[width=0.55\textwidth]
    {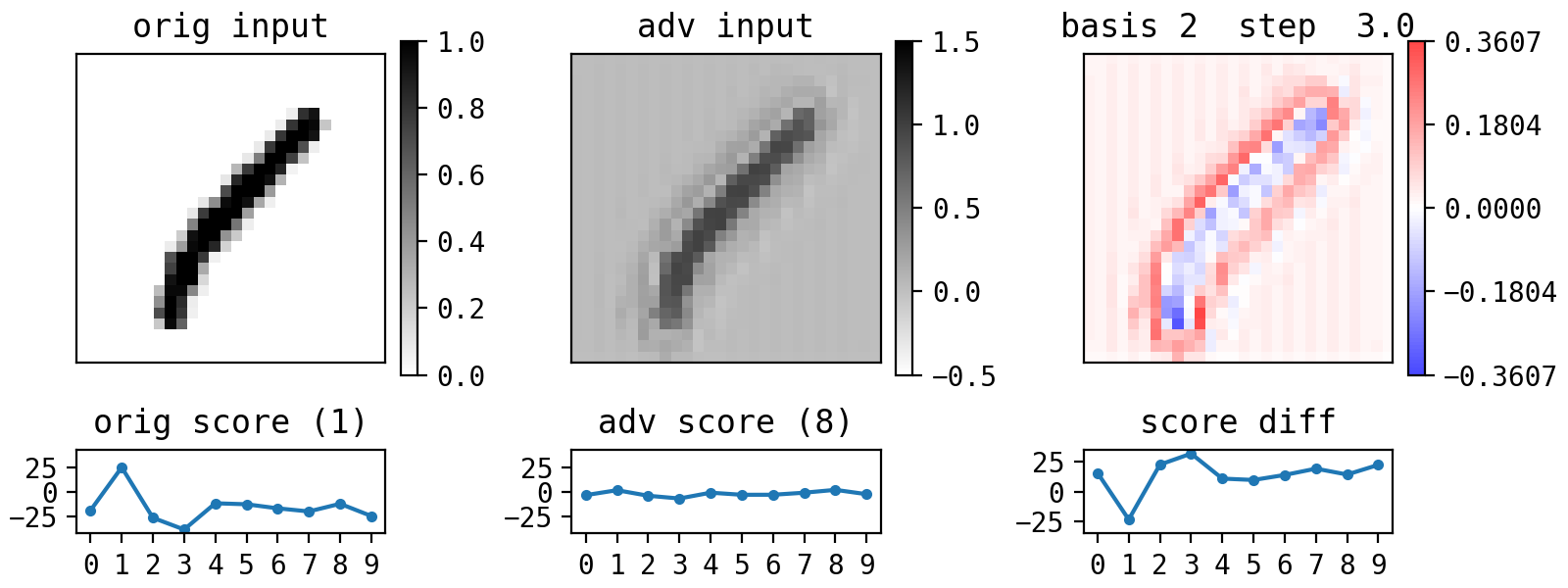}
  \includegraphics[width=0.55\textwidth]
    {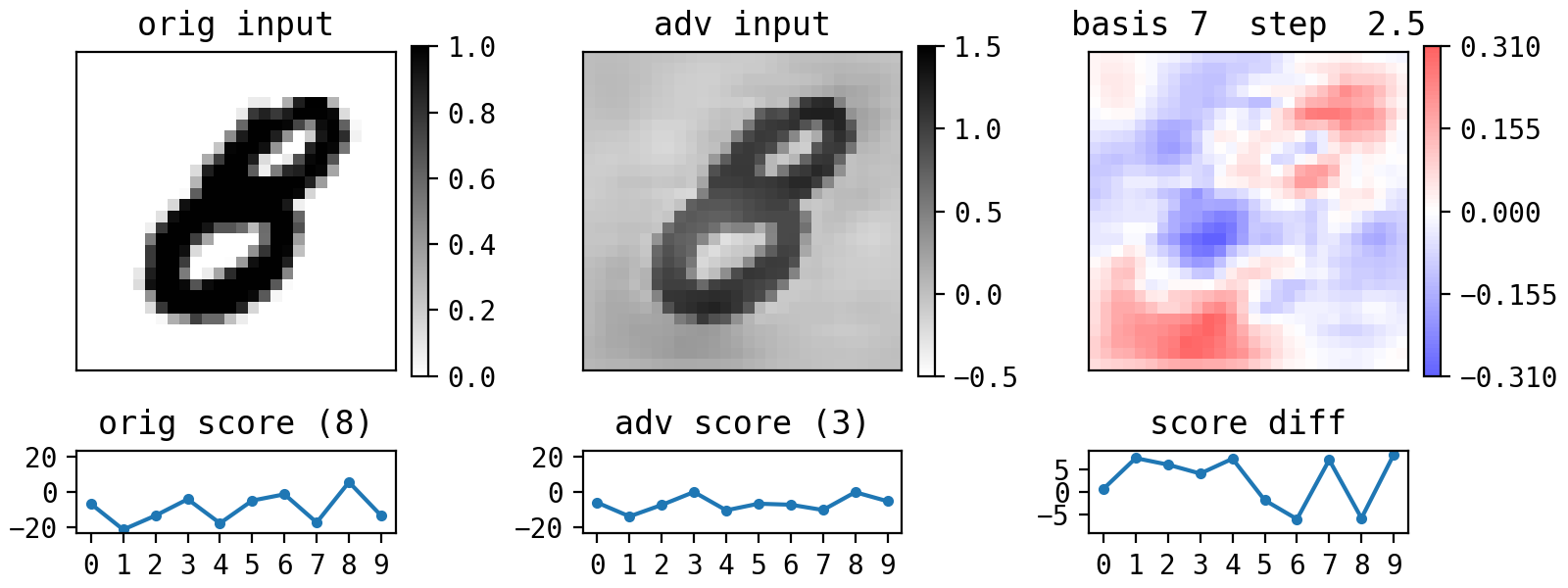}
  \includegraphics[width=0.55\textwidth]
    {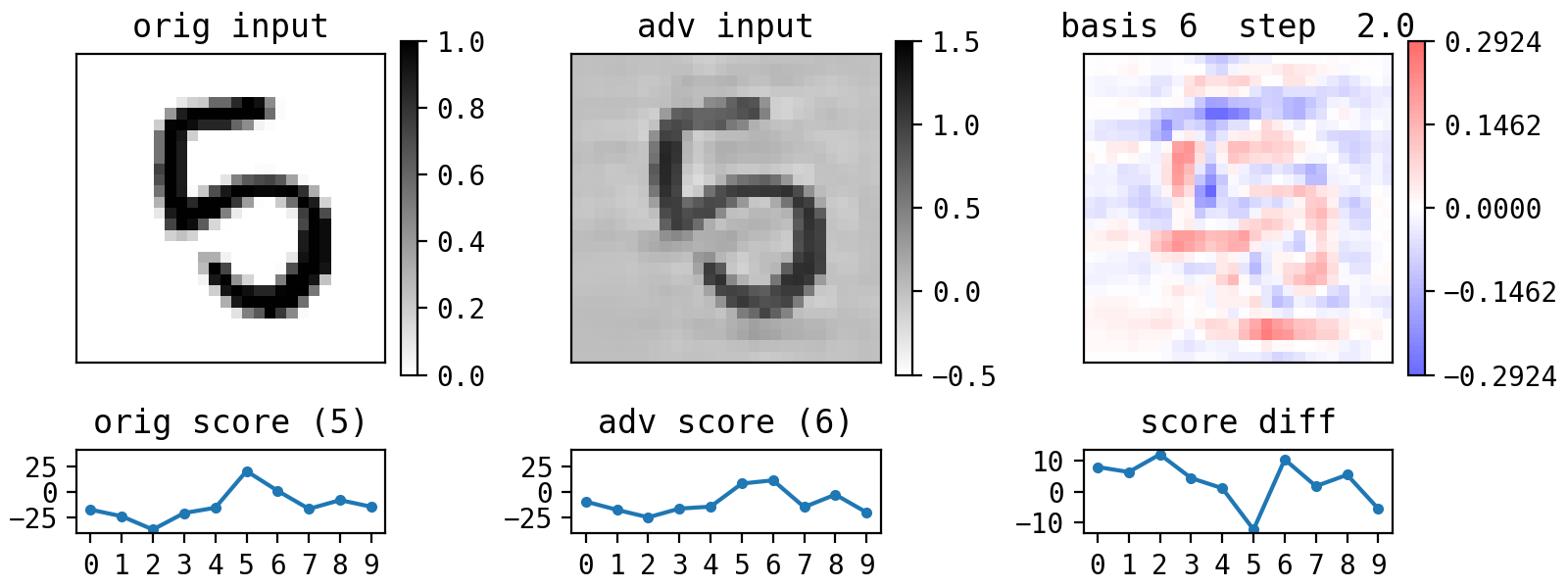}
  \includegraphics[width=0.55\textwidth]
    {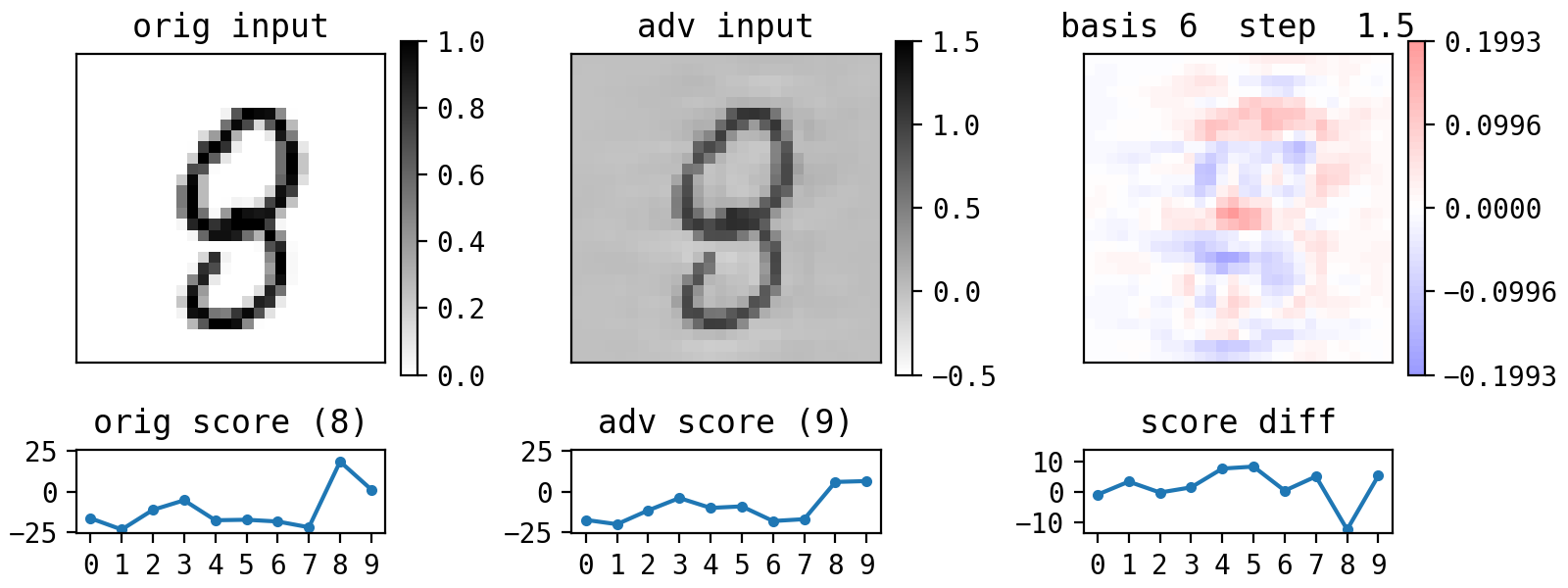}
  \includegraphics[width=0.55\textwidth]
    {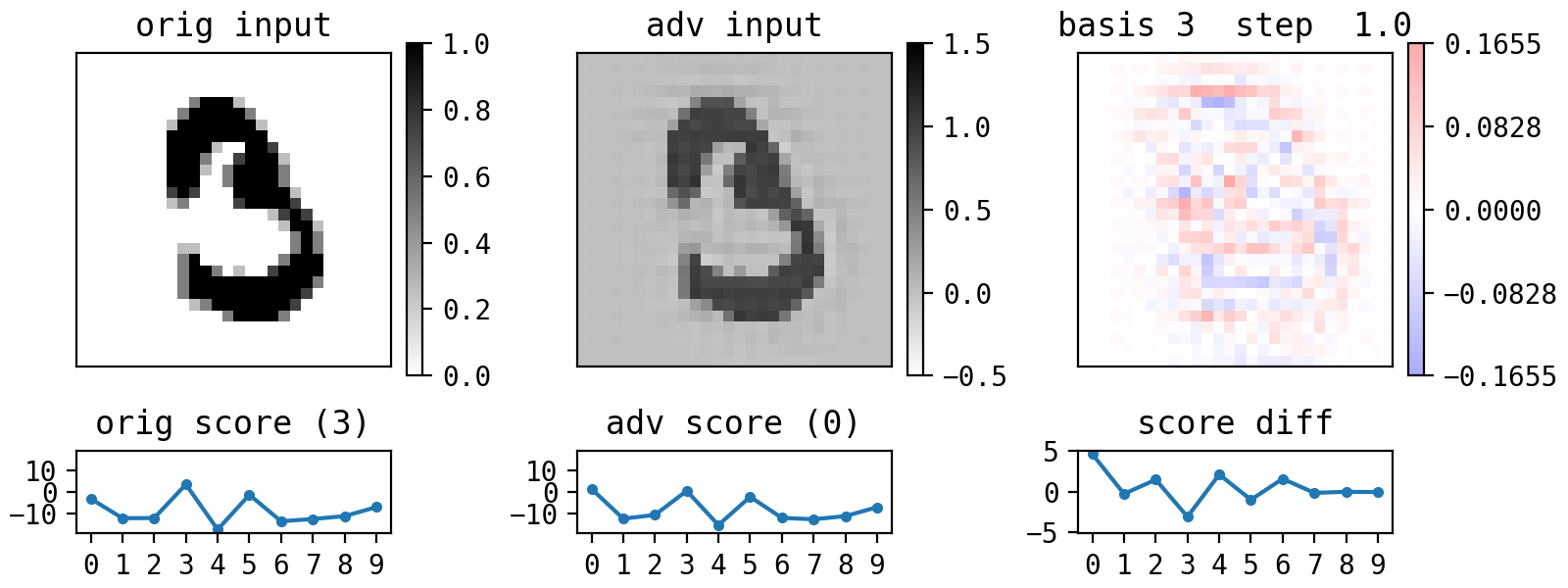}
  \includegraphics[width=0.55\textwidth]
    {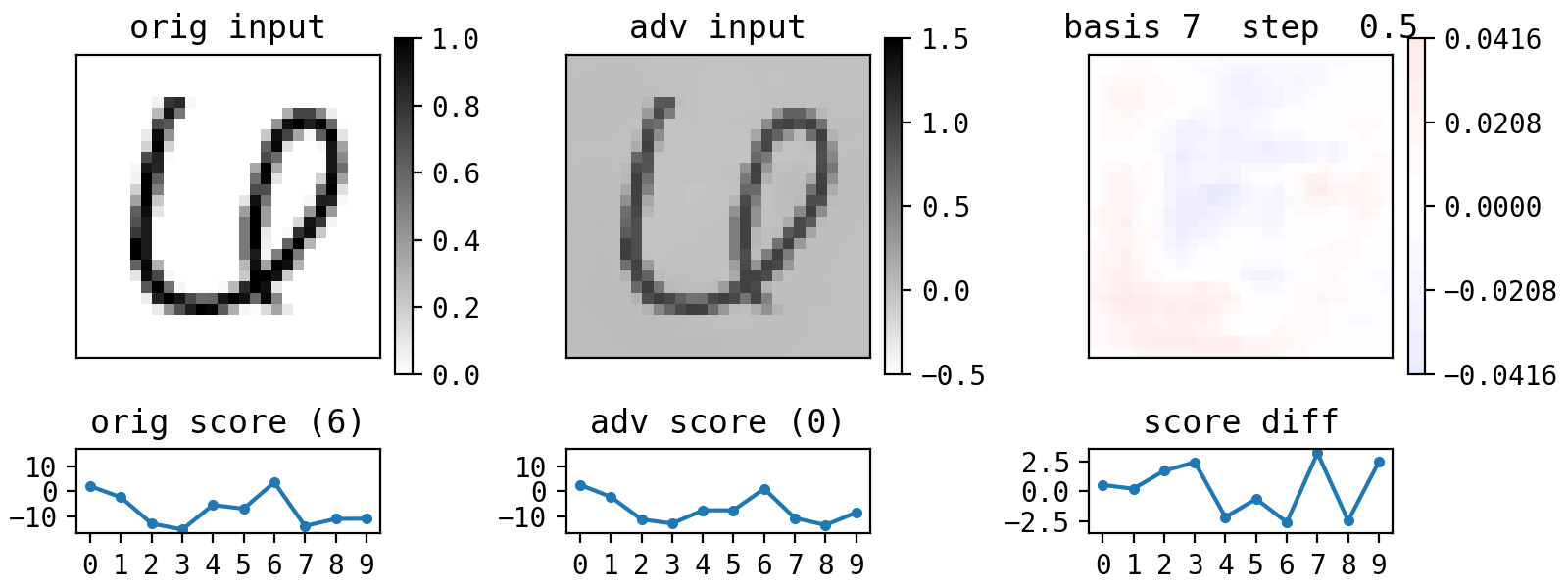}
  \caption{LRHE basis attack results. Six adversarial examples found for varying
  step-size $\alpha$. The original image and correctly predicted label is shown,
  alongside the perturbed input and the perturbation. The corresponding logit
  scores and the changes are also shown below the examples.}
  \label{fig:lrhe_basis_attack}
\end{figure}

This LRHE basis attack had a success rate of 7.79\% as shown in
Table~\ref{tab:lrhe-attack}, showing that these basis functions are indeed
related to directions in the input space that the model is sensitive with
respect to. Larger $\alpha$'s tend to yield more examples, which is to be
expected. To ascertain that this is a significant success rate, we conduct the
same experiment with $\alpha \le 3.0$ and random vectors in place of 
$\psi_\ell$, and find a median success rate of 1.88\% (max 1.96\%, min 1.81\%,
five different trials).

This shows that LRHE input basis functions span a linear space of perturbations
that are significantly adversarial. On the other hand, this attack result is not
as strong as {\tt AutoAttack} which had a success rate of 28.01\% for the range
$\alpha \le 1.0$ (see Table~\ref{tab:robust} below).

\subsection{Projection to the orthogonal complement of LRHE basis}  
\begin{figure}
  \includegraphics[width=1.0\textwidth] {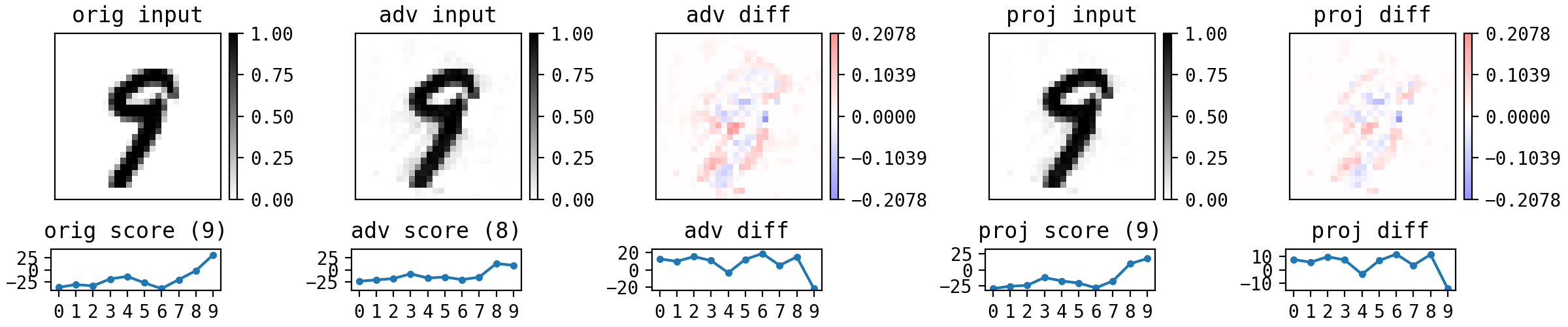}
  \includegraphics[width=1.0\textwidth] {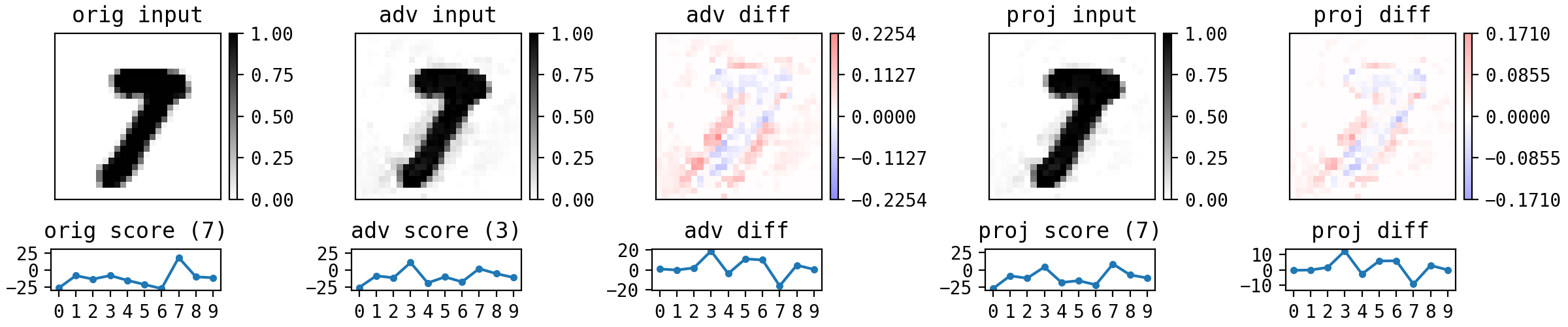}
  \includegraphics[width=1.0\textwidth] {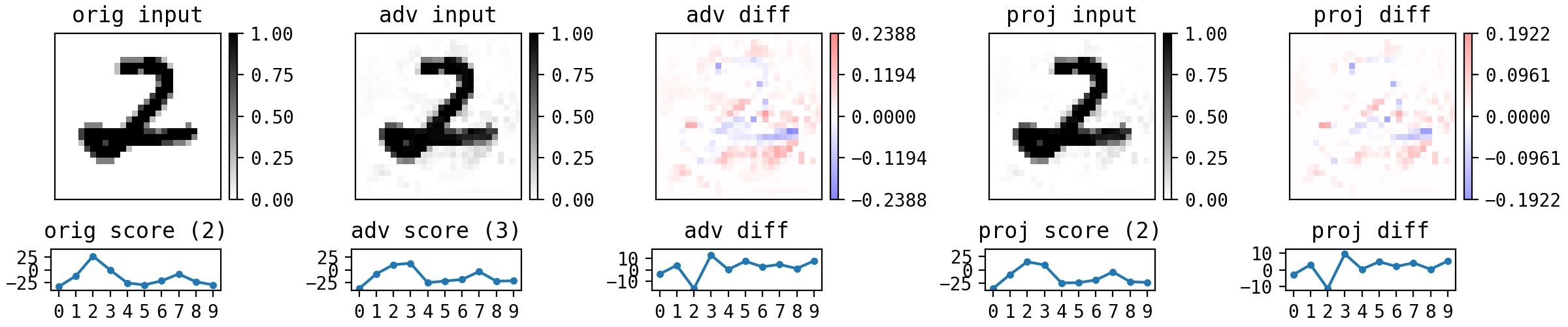}
  \includegraphics[width=1.0\textwidth] {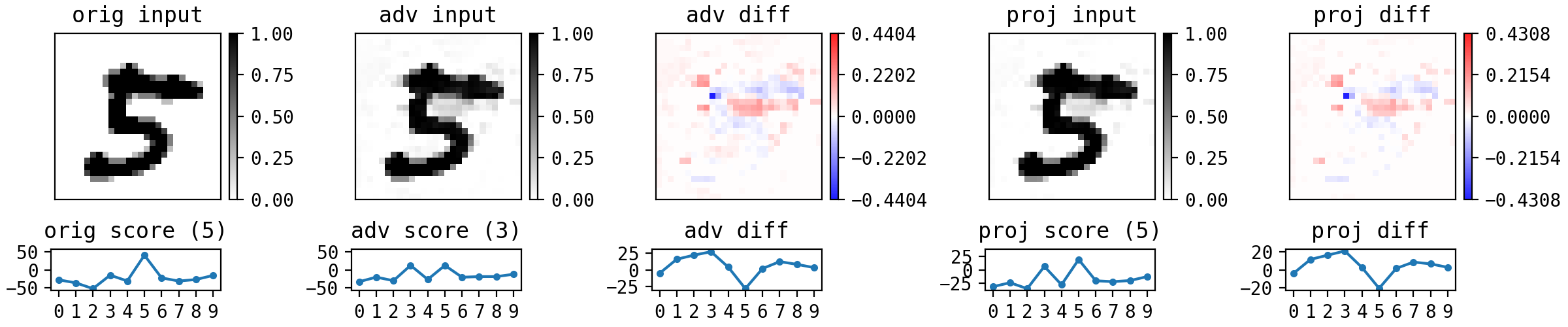}
  \caption{Orthogonal complement projection of adversarial examples leads to
  corrections. Original test image, the adversarial example, and the projected
  example are shown, along with the differences. Output logit scores and
  corresponding changes are also shown.}
  \label{fig:proj}
\end{figure}

Next, we check if the LRHE input basis $(\psi_\ell)_\ell$ can be used to remove
the adversarial effects from examples discovered through {\tt AutoAttack}. We
form an orthogonal matrix by forming the matrix $\Psi$ whose columns are made up
of the LRHE input basis, then taking the QR factorization to obtain the matrix
$\widehat{\Psi}$ with orthonormal columns,
\begin{equation}
  \Psi := [\psi_1 \mid \cdots \mid \psi_{L-1}],
  \quad
  \widehat{\Psi} R = \Psi,
  \quad
  \widehat{\Psi}^\top \widehat{\Psi} = I.
\end{equation}
For each adversarial example $\delta x$, we compute the projected version
$\delta x_{\text{proj}}$
\begin{equation}
  \delta x_{\text{proj}}
  =
  \text{clip}_{[0,1]}(
        (I - \widehat{\Psi} \widehat{\Psi}^\top) \cdot \delta x
      ),
\end{equation}
then check if these examples are still adversarial. After projections, a
majority of the adversarial examples no longer switched the predicted labels,
that is, the projected perturbations were no longer adversarial.  For the
projected examples, the adversarial success rate fell to 9.20\% from 28.01\%,
about a third. To check that this correction is significant, we conduct the same
experiment with a random orthogonal basis, and find median success rate 25.42\%
(max.\ 25.45\%, min.\ 25.34\%, five different trials). A sample of these
projection test results is shown in Figure~\ref{fig:proj}. Note that a similar
test was performed for the regression task in \cite{lrhe2024}.

\section{Adversarial training results} \label{sec:results}

\subsection{Training heuristics} The regularized training objective
\eqref{eq:obj} requires more fine-tuning compared to the standard objective
involving only the misfit loss.  Straightforward Adam optimizer with standard
fixed choices for $\lambda$ and $\gamma$ resulted in improved adversarial
robustness but test accuracy was 1--2\% lower than that of the original model. 
As the regularized problem has new hyper-parameters $\lambda$ and $\gamma$, we
tune them to achieve better results.  As for the hyper-parameter $\gamma$, it is
sampled randomly from a uniform distribution at each iteration, $\gamma_k \sim
\cU[-1, 1].$ Note that $\gamma$ is not to be confused with the budget $\eps$; it
is a dimensionless scale on $\phi_\ell$ and not a perturbation radius.  The
regularization parameter $\lambda$ is updated at each Adam iteration to be
adjusted from its base value $\lambda_{\textrm{base}} = \texttt{1e-2}$, that is,
at iteration $k$,
\begin{equation}
  \lambda_k 
  := 
  \frac{\cL(f_{\theta_k}(x), y)}{\cR(\theta_k, x ; \gamma_k)}
  \cdot
  \lambda_{\textrm{base}}
\end{equation}
so that the two terms are roughly at the same scale during training. This
adjustment is done before the backward pass computation.

We also deploy a fixed decay rate scheduler {\tt MultiplicativeLR} implemented
in PyTorch, which reduces the learning rate by {\tt factor}={\tt 0.9} every two
epochs. We run up to 70 epochs of Adam optimizer, and select the model with the
best validation accuracy amoung 25 different random initializations.  The
trained model achieves test accuracy of 99.53\%, on par with the original model.

To provide a comparison with adversarial examples with bounded
$\ell^\infty$-norms we set $\eps=0.05$, the level at which the adversarial
accuracy of the original model is similar to our setting. The threat model is
again {\tt AutoAttack}, which yielded stronger attack results than our own PGD
attack which had the configuration 25 runs of PGD attack with 500 iterations
with step-size 0.01 plus 25 runs of PGD attack with 500 iterations and step-size
0.005.

\begin{figure}
  \centering
  \includegraphics[width=1.0\textwidth]{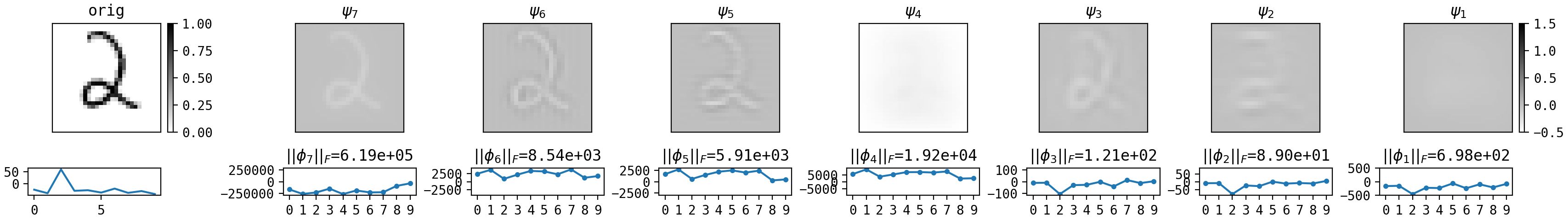}
  \includegraphics[width=1.0\textwidth]{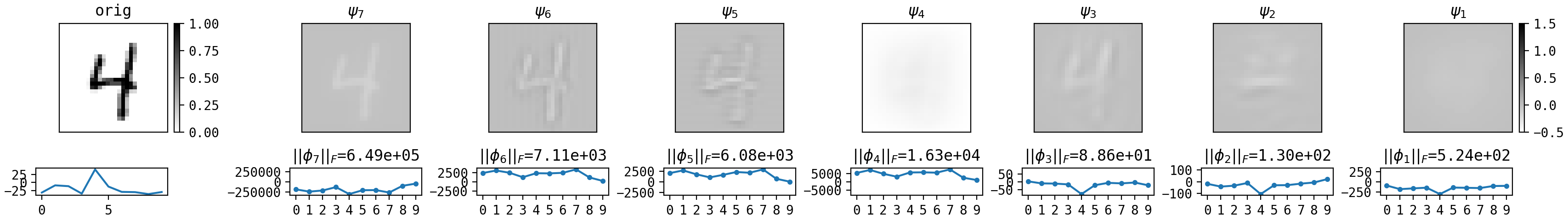}
  \vskip2pt
  \includegraphics[width=1.0\textwidth]{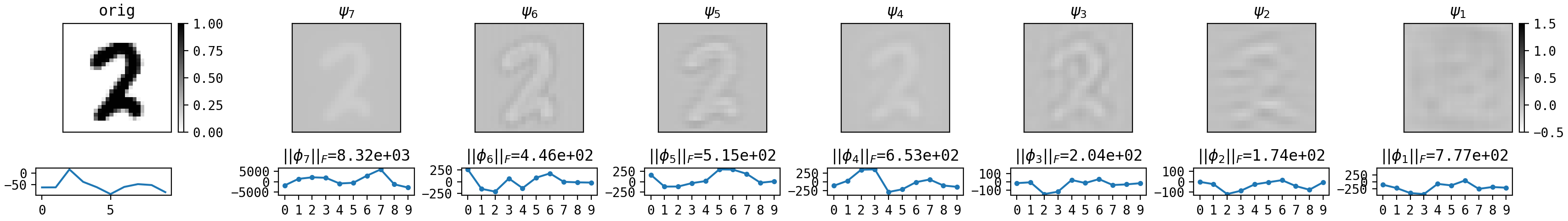}
  \includegraphics[width=1.0\textwidth]{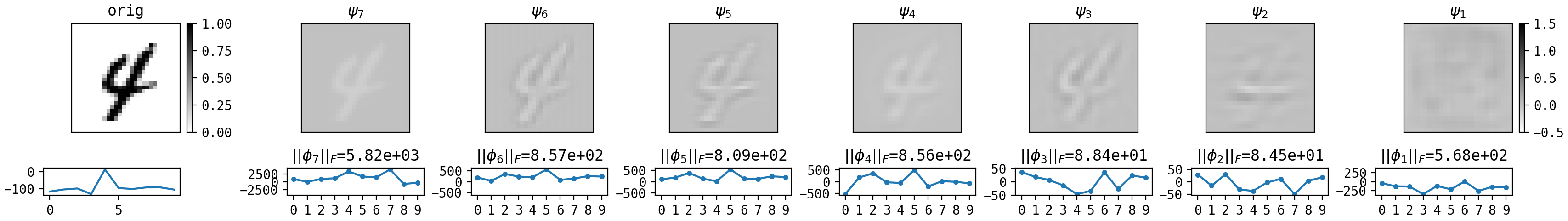}
  \caption{LRHE input and output bases for original and adversarially trained
  models. Top two examples are from the LRHE of the original model,
  and bottom two examples from the adversarially trained model.}
  \label{fig:basis_orig_reg}
\end{figure}

\subsection{Adversarial robustness results} 
To illustrate the effect of the adversarial training, we compare the input and
output basis of the clean model and adversarially trained model.  The norm of
the output basis is significantly smaller in the case of the regularized model,
by two orders of magnitude; see Figure~\ref{fig:basis_orig_reg} for a sample of
LRHE input and output basis.

\begin{table}
    \begin{tabular}{rr|llll}
    \hline
                    & & Original & LRHE-AT    &  PGD-40-AT  & PGD-3-AT
    \\ \hline
    & Test & 99.48 & 99.53  & 99.46 & 99.43
    \\
    \hline
    \multirow{6}{*}{\rotatebox{90}{$\ell^2$-rel. adv.}}
    & $\eps=0.002$ &99.45 (\z0.03) & 99.48 (\z0.05)& 99.40 (\z0.06) & 99.34 (\z0.09)
    \\ 
    & $\eps=0.004$ &99.42 (\z0.06) & 99.40 (\z0.13)& 99.36 (\z0.10) & 99.31 (\z0.12)
    \\ 
    & $\eps=0.006$ &99.34 (\z0.14) & 99.38 (\z0.15)& 99.28 (\z0.18) & 99.26 (\z0.17)
    \\ 
    & $\eps=0.008$ &99.25 (\z0.23) & 99.31 (\z0.22)& 99.25 (\z0.21) & 99.25 (\z0.18)
    \\ 
    & $\eps=0.010$ &99.17 (\z0.31) & 99.27 (\z0.26)& 99.20 (\z0.26) & 99.19 (\z0.24)
    \\ 
    & $\eps=0.012$ &99.10 (\z0.38) & 99.20 (\z0.33)& 99.16 (\z0.30) & 99.17 (\z0.26)
    \\ 
    & $\eps=0.014$ &98.96 (\z0.52) & 99.13 (\z0.40)& 99.14 (\z0.32) & 99.10 (\z0.33)
    \\ 
    & $\eps=0.016$ &98.84 (\z0.64) & 99.09 (\z0.44)& 99.12 (\z0.34) & 99.06 (\z0.37)
    \\ 
    & $\eps=0.018$ &98.71 (\z0.77) & 99.01 (\z0.52)& 99.08 (\z0.38) & 98.99 (\z0.44)
    \\ 
    & $\eps=0.020$ &98.63 (\z0.85) & 98.97 (\z0.56)& 99.02 (\z0.44) & 98.96 (\z0.47)
    \\ 
    & $\eps=0.025$ &98.28 (\z1.20) & 98.83 (\z0.70)& 98.89 (\z0.57) & 98.86 (\z0.57)
    \\ 
    & $\eps=0.050$ &95.47 (\z4.20) & 97.50 (\z2.03)& 98.10 (\z1.36) & 97.93 (\z1.50)
    \\ 
    & $\eps=0.075$ &88.40  (11.08) & 94.94 (\z4.59)& 96.64 (\z2.82) & 96.15 (\z3.28)
    \\ 
    & $\eps=0.100$ &71.47  (28.01) & 88.47 (11.06) & 93.69 (\z5.77) & 92.75 (\z6.68)
    \\ 
    & $\eps=0.125$ &44.58  (54.90) & 77.57 (21.96) & 87.58 (11.88) & 85.68 (13.75)
    \\ 
    & $\eps=0.150$ &17.54  (81.94) & 60.33 (39.20) & 76.03 (23.43) & 72.53 (26.90)
    \\ 
    \hline
    \multirow{6}{*}{\rotatebox{90}{$\ell^\infty$ adv.}}
    & $\eps=0.010$ &98.91  (\z0.57) & 99.12 (\z0.41) & 99.17 (\z0.29)& 99.15 (\z0.28)
    \\                                                       
    & $\eps=0.025$ &97.22  (\z2.26) & 98.50 (\z1.03) & 98.73 (\z0.73)& 98.71 (\z0.72)
    \\                                                       
    & $\eps=0.050$ &90.32  (\z9.16) & 95.46 (\z4.07) & 97.44 (\z2.02)  & 97.34 (\z2.09)
    \\                                                       
    & $\eps=0.075$ &66.56  (32.92) & 87.55 (11.98)  & 94.91 (\z4.55)  & 93.99 (\z5.44)
    \\                                                       
    & $\eps=0.100$ &25.00  (74.48) & 70.05 (29.48)  & 88.67 (10.79) & 86.47 (12.96)
    \\                                                       
    & $\eps=0.125$ &\z2.52 (96.96) & 40.73 (58.80)  & 74.61 (24.85) & 68.71 (30.72)
    \\
    & $\eps=0.150$ &\z0.00 (99.48) & 12.57 (86.96) & 47.76 (51.70) & 38.42 (61.01)
    \\
    \hline
  \end{tabular}
  \caption{Adversarial accuracy (percentage) under relative $\ell^2$
  perturbations ($\Norm{\delta x}{2} \le \eps\|x\|_2$; with $\|x\|_2 \approx
  9.2$ on MNIST, $\eps = 0.150$ corresponds to an absolute radius $\| \delta
  x\|_2$ of $\approx 1.4$) and under $\ell^\infty$ perturbations. All defended
  models were trained at $\eps = 0.075$ in the relative $\ell^2$ norm; the
  $\ell^\infty$ block therefore measures cross-norm transfer. Parenthesized
  values give accuracy lost relative to each model's own clean accuracy.}
  \label{tab:robust}
\end{table}

The adversarial training results are shown in Table~\ref{tab:robust}.

\subsubsection{Clean accuracy}

Table~\ref{tab:robust} reports clean test accuracy alongside adversarial
accuracy under both norms. The four models reach $99.48\%$ for the unregularized
baseline, $99.53\%$ for the LRHE-regularized model, $99.46\%$ for $40$-step and
$99.43\%$ for 3-step adversarial training. These are separated by at most ten
test examples, so the ordering is inconclusive.  The loss of clean accuracy that
adversarial training is sometimes reported to incur \cite{tsipras2019robustness}
is not visible at this scale, though this is potentially because MNIST clean
accuracy is saturated. All four models attain clean accuracy above $99.4\%$, and
we therefore confine the comparison to adversarial accuracy.

\subsubsection{The small-budget regime}

At small perturbation budgets the regularized model matches both PGD
configurations. Under relative $\ell^2$ perturbations at $\eps = 0.010$ it
retains $99.27\%$ against $99.20\%$ for $40$-step and $99.19\%$ for 3-step
training; at $\eps = 0.020$ the three stand at $98.97\%$, $99.02\%$ and
$98.96\%$. Across the range $\eps \le 0.020$ the three defended models are on
par, differences among them between one and fourteen test examples.  This parity
is of particular interest: a regularizer that takes no derivative with respect
to the input is indistinguishable here from adversarial training that takes 40
per update.

\subsubsection{Larger budgets}

This parity does not extend to larger budgets.  Under relative $\ell^2$
perturbations the PGD-40-AT model moves ahead between $\eps = 0.012$ and $\eps =
0.014$, and the PGD-3-AT model between $\eps = 0.020$ and $\eps = 0.025$; past
these points PGD-AT is increasingly robust by comparison, reaching a margin of
$15.70$ percentage points over the regularized model at $\eps = 0.150$. Under
$\ell^\infty$ perturbations multi-step training is ahead at every budget we
tabulate, and the margin at $\eps = 0.150$ is $35.19$ percentage points.  To
summarize, the regularizer matches PGD-AT for $\eps \le 0.02$ in the relative
$\ell^2$ norm, and is weaker beyond that range, as well as under $\ell^\infty$
at budgets $\eps \ge 0.01$.

This behavior is consistent with the approximation on which the method relies.
The expansion truncates the ReLU representation at rank one, that is, once the
vector $v_\ell$ is fixed it is first-order accurate in a cone about the input
whose extent is determined by the neglected terms; as the perturbation budget
grows, the adversary is increasingly free to leave the region in which the
retained directions describe the network's local action. Degradation at large
$\eps$ is thus predicted by the construction, and suggests that the usable range
should extend with improvement of the approximation \eqref{eq:expand}.

\subsubsection{Asymmetry between the two norms}

The steeper decline under $\ell^\infty$ merits comment. The subspace identified
by the expansion is defined by orthogonal projections and is intrinsically an
$\ell^2$ object. A fixed $\ell^\infty$ budget on MNIST admits perturbations of
$\ell^2$ norm up to $\sqrt{\ninput} \eps = 28\eps$, so the $\ell^\infty$ rows of
Table~\ref{tab:robust} probe substantially larger $\ell^2$ displacements than
the corresponding relative-$\ell^2$ rows. The asymmetry is therefore expected,
and the two blocks of the table should not be read as directly comparable.

\begin{table}
  \begin{tabular}{r|l}
    \hline
                              & LRHE-AT 
    \\ \hline
    Test accuracy             &   99.53    
    \\\hline
    White-box adv. acc.       &   94.94 (4.59) 
    \\
    \hline
    Transfer from second LRHE-AT model   &  97.28 (2.72) 
    \\ \hline
    Transfer from original model &  98.47 (1.53) 
    \\ \hline
  \end{tabular}
  \caption{Black-box attack for the adversarially trained model.
  Adversarial accuracy (percentage) computed for $\ell^2$-relative $\eps=0.075$.
  }
  \label{tab:blackbox}
\end{table}

\subsubsection{Black-box attacks} We adversarially train a second model with a
different random initialization, apply the same white-box {\tt AutoAttack} to
the new model to find adversarial examples, then use these examples to attack
our adversarially trained model, in a black-box attack. The adversarial accuracy
of our adversarially trained model with respect to white-box attack was
$94.94\%$, whereas black-box attacks using adversarial examples from the second
model yield an adversarial accuracy of $97.28\%$, and black-box attacks using
examples from the original (clean) model yield $98.47\%$. The results are shown
in Table~\ref{tab:blackbox}.

\begin{table}

  \begin{tabular}{r|cccccc}
    \hline
    PGD steps & 1 & 2 & 5 & 10 & 20 & 40
    \\ \hline
    Wall time (s) & 7.8 & 10.6 & 18.8 & 32.3 & 58.0 & 110.5 
    \\ \hline
  \end{tabular}
  
  \caption{Wall time per epoch during PGD adversarial training. 
          Batch size is 30.  The wall time was measured using standard PyTorch
          implementation on NVIDIA GeForce RTX 3090 with 24GB of memory. 
          }
\end{table}

\subsubsection{Computational cost}

Wall-clock time per epoch for PGD adversarial training is linear in the number
of inner steps,
\begin{equation}
  T(K) \approx 5.4 + 2.65 K \quad \text{seconds},
  \label{eq:pgdtime}
\end{equation}
at batch size $30$, the fixed term reflecting the cost of a standard training
epoch and the marginal term the cost of one input-gradient computation over the
training set. LRHE-regularized training requires $12.7$s/epoch under identical
conditions, equivalent to $K \approx 2.8$ PGD steps and an $8.7\times$ reduction
relative to the $40$-step configuration of Table~\ref{tab:robust}. It is also
below the $13.4$s that \eqref{eq:pgdtime} predicts for 3-step training, so the
cost-matched comparison in Table~\ref{tab:robust} is made against a baseline
that is if anything slightly more expensive than our method. Single-step methods
such as those of \cite{wong2020fast} and \cite{shafahi2019free} remain cheaper
in absolute terms; our claim is not to be the cheapest available defense but to
obtain robustness at a fraction of the cost of multi-step training without
differentiating with respect to the input at any point.  This is a
constant-factor trade rather than an asymptotic improvement. LRHE does not
eliminate the extra computation adversarial training requires: Where PGD
performs $K$ sequential forward-backward passes differentiating with respect to
the input, LRHE performs a single parameter-gradient computation over a forward
map that is $L$ times more expensive. The comparison therefore has a critical
point: LRHE is cheaper than $K$-step PGD training for $K \ge 3$ and more
expensive below it. Since robust accuracy on MNIST is conventionally obtained
with $K=40$, the regime of interest lies well above this point.

The two costs also differ in structure, and not only in size. The $L-1$ partial
products in \eqref{eq:output_basis} are mutually independent, and they share
suffixes: $\phi_\ell$ and $\phi_{\ell'}$ pass through the same affine maps above
$\max(\ell,\ell')$. They may therefore be accumulated in a single sweep down the
layers, adding the correction $\sigma z_\ell - z_\ell$ at each stage before
applying the next affine map, rather than being formed one at a time. The $K$
steps of projected gradient ascent admit no such reorganization, since each step
requires the perturbation produced by the one before it. The timings above were
obtained without exploiting this, and are in that respect conservative with
respect to the regularizer.

\subsubsection{Limitations}
Two limitations of this evaluation should be stated plainly. First, both
defended models were trained at $\eps = 0.075$ with the relative $\ell^2$-norm,
and we evaluate over budgets up to $\eps = 0.150$; we do not report results at
the $\eps = 0.3$ $\ell^\infty$ budget conventional for MNIST, at which neither
model retains useful accuracy. Second, all results are on MNIST, and we refrain
from claiming that the accuracy of the approximation required to capture the
adversarially relevant subspace, or the budget range over which the
approximation remains accurate, transfers to higher-resolution data.

\section{Conclusion}

We have argued that for adversarial examples of small relative $\ell^2$ norm,
the directions along which a network is most sensitive need not be searched for.
They are already determined by the activation pattern, and the expansion of
\cite{lrhe2024} makes them available as a byproduct of the forward pass.  The
consequence for adversarial training is that the inner maximization of the
min-max formulation can be replaced by a penalty on a fixed set of directions
recomputed at each step, so that no derivative with respect to the input is
taken at any point during training.  The main contribution of this work is to
establish that input-gradient-free adversarial training is possible. The other
methods that dispense with the inner maximization invariably measure the local
geometry by differentiating with respect to the input; the construction here
shows that this is not required.

The following observations are made regarding the proposed construction for the
MNIST setting. The subspace spanned by the input basis is where much of the
adversarial perturbations lie: it has dimension seven against an input dimension
of $784$, and projecting perturbations onto its orthogonal complement removes
roughly two thirds of the adversarial effect, against almost none for a random
subspace of the same dimension. The resulting training procedure costs the
equivalent of $2.8$ PGD steps per epoch, less than 3-step adversarial training
and a factor of $8.7$ below 40-step. Within the small budget range for which the
expansion is descriptive, the models it produces match both, the 3-step robust
accuracy up to $\eps \le 0.02$ and the 40-step robust accuracy up to $\eps \le
0.012$; beyond that range the multi-step training pulls ahead, for the reasons
discussed below.  Still, we regard the small-budget regime as the setting in
which the original motivation for studying adversarial examples, that the
perturbation be imperceptible, is still intact.  Finally, the robustness so
obtained does not appear to rest on gradient masking: adversarial examples
transferred from an independently trained model, and from the undefended model,
are considerably less effective against the regularized network than examples
constructed against it directly, which is the ordering a genuine defense is
expected to show \cite{athalye2018obfuscated}.

The limits of the method follow from the same source as its efficiency. The
reflection vectors are fixed by the activation pattern at the clean input, so
the description they provide is local, and it ceases to hold once a perturbation
is large enough to cross activation boundaries in quantity.  Multi-step
adversarial training is under no such restriction, and at large budgets the
difference is substantial.  Our directions are moreover independent of the
label: they record where the map amplifies, not where the decision boundary
lies, and the two coincide only approximately.

Several future directions follow naturally. The most immediate is a wider
empirical studies: results on datasets where clean accuracy is not saturated,
and where the robustness-accuracy tradeoff is pronounced enough that the absence
of a drop would carry weight. Comprehensive replication across random seeds is
needed before the margins reported here at individual budgets can be regarded as
settled.

A second direction concerns architecture. The two operations underlying the
construction, taking the difference between pre- and post-activation states and
propagating it through the remaining affine maps, do not depend on the network
being feedforward. They require only that the nonlinearity be separable from the
affine structure at a given input, which holds for attention layers as well,
where the attention matrix computed in the forward pass plays the role that the
activation pattern plays here. Whether the resulting subspace carries the same
relation to adversarial perturbations in that setting is open.

Finally, the input basis has uses beyond the training procedure studied here.
It is cheap to compute, it requires no differentiation, and in the absence of
adversarial examples it identifies the input directions to which the model
responds most strongly. That makes it a candidate tool for sensitivity analysis
of trained models generally, independently of any robustness objective. 

\subsection*{Code availability}
The code used to generate the results above are publicly accessible in the code
repository \cite{lrhe-at-code}.

\bibliographystyle{siamplain}

\begin{thebibliography}{10}

\bibitem{andriushchenko2020square}
{\sc M.~Andriushchenko, F.~Croce, N.~Flammarion, and M.~Hein}, {\em Square attack: A query-efficient black-box adversarial attack via random search}, in European Conference on Computer Vision (ECCV), 2020.

\bibitem{andriushchenko2020understanding}
{\sc M.~Andriushchenko and N.~Flammarion}, {\em Understanding and improving fast adversarial training}, in Advances in Neural Information Processing Systems (NeurIPS), 2020.

\bibitem{athalye2018obfuscated}
{\sc A.~Athalye, N.~Carlini, and D.~Wagner}, {\em Obfuscated gradients give a false sense of security: Circumventing defenses to adversarial examples}, in Proceedings of the 35th International Conference on Machine Learning (ICML), vol.~80 of Proceedings of Machine Learning Research, 2018, pp.~274--283.

\bibitem{carlini2019evaluating}
{\sc N.~Carlini, A.~Athalye, N.~Papernot, W.~Brendel, J.~Rauber, D.~Tsipras, I.~Goodfellow, A.~Madry, and A.~Kurakin}, {\em On evaluating adversarial robustness}, arXiv preprint arXiv:1902.06705,  (2019).

\bibitem{cisse2017parseval}
{\sc M.~Cisse, P.~Bojanowski, E.~Grave, Y.~Dauphin, and N.~Usunier}, {\em Parseval networks: Improving robustness to adversarial examples}, in Proceedings of the 34th International Conference on Machine Learning (ICML), vol.~70 of Proceedings of Machine Learning Research, 2017.

\bibitem{croce2020minimally}
{\sc F.~Croce and M.~Hein}, {\em Minimally distorted adversarial examples with a fast adaptive boundary attack}, in Proceedings of the 37th International Conference on Machine Learning (ICML), vol.~119 of Proceedings of Machine Learning Research, 2020.

\bibitem{croce2020reliable}
{\sc F.~Croce and M.~Hein}, {\em Reliable evaluation of adversarial robustness with an ensemble of diverse parameter-free attacks}, in Proceedings of the 37th International Conference on Machine Learning (ICML), vol.~119 of Proceedings of Machine Learning Research, 2020, pp.~2206--2216.

\bibitem{fazlyab2019efficient}
{\sc M.~Fazlyab, A.~Robey, H.~Hassani, M.~Morari, and G.~J. Pappas}, {\em Efficient and accurate estimation of {Lipschitz} constants for deep neural networks}, in Advances in Neural Information Processing Systems (NeurIPS), 2019.

\bibitem{goodfellow2015explaining}
{\sc I.~J. Goodfellow, J.~Shlens, and C.~Szegedy}, {\em Explaining and harnessing adversarial examples}, in International Conference on Learning Representations (ICLR), 2015.

\bibitem{hendrycks2021natural}
{\sc D.~Hendrycks, K.~Zhao, S.~Basart, J.~Steinhardt, and D.~Song}, {\em Natural adversarial examples}, in IEEE/CVF Conference on Computer Vision and Pattern Recognition (CVPR), 2021, pp.~15262--15271.

\bibitem{hoffman2019robust}
{\sc J.~Hoffman, D.~A. Roberts, and S.~Yaida}, {\em Robust learning with jacobian regularization}, arXiv preprint arXiv:1908.02729,  (2019).

\bibitem{Kingma2015AdamAM}
{\sc D.~P. Kingma and J.~Ba}, {\em Adam: A method for stochastic optimization}, in International Conference on Learning Representations (ICLR), 2015.

\bibitem{li2019preventing}
{\sc Q.~Li, S.~Haque, C.~Anil, J.~Lucas, R.~Grosse, and J.-H. Jacobsen}, {\em Preventing gradient attenuation in {Lipschitz} constrained convolutional networks}, in Advances in Neural Information Processing Systems (NeurIPS), 2019.

\bibitem{li2022subspace}
{\sc T.~Li, Y.~Wu, S.~Chen, K.~Fang, and X.~Huang}, {\em Subspace adversarial training}, in IEEE/CVF Conference on Computer Vision and Pattern Recognition (CVPR), 2022.

\bibitem{madry2018towards}
{\sc A.~Madry, A.~Makelov, L.~Schmidt, D.~Tsipras, and A.~Vladu}, {\em Towards deep learning models resistant to adversarial attacks}, in International Conference on Learning Representations (ICLR), 2018.

\bibitem{miyato2018spectral}
{\sc T.~Miyato, T.~Kataoka, M.~Koyama, and Y.~Yoshida}, {\em Spectral normalization for generative adversarial networks}, in International Conference on Learning Representations (ICLR), 2018.

\bibitem{moosavi2019robustness}
{\sc S.-M. Moosavi-Dezfooli, A.~Fawzi, J.~Uesato, and P.~Frossard}, {\em Robustness via curvature regularization, and vice versa}, in IEEE/CVF Conference on Computer Vision and Pattern Recognition (CVPR), 2019.

\bibitem{qin2019adversarial}
{\sc C.~Qin, J.~Martens, S.~Gowal, D.~Krishnan, K.~Dvijotham, A.~Fawzi, S.~De, R.~Stanforth, and P.~Kohli}, {\em Adversarial robustness through local linearization}, in Advances in Neural Information Processing Systems (NeurIPS), 2019.

\bibitem{lrhe-at-code}
{\sc T.C.~Johnson, D.~Rim}, {\em dsrim/lrhe-at: Code repository}, Aug. 2026, \href{http://dx.doi.org/10.5281/zenodo.22119131}{doi:\nolinkurl{10.5281/zenodo.22119131}}.

\bibitem{rim2022tsunami}
{\sc D.~Rim, R.~Baraldi, C.~M. Liu, R.~J. LeVeque, and K.~Terada}, {\em Tsunami early warning from global navigation satellite system data using convolutional neural networks}, Geophysical Research Letters, 49 (2022), p.~e2022GL099511, \href{http://dx.doi.org/10.1029/2022GL099511}{doi:\nolinkurl{10.1029/2022GL099511}}.

\bibitem{lrhe2024}
{\sc D.~Rim, S.~Suri, S.~Hong, K.~Lee, and R.~J. LeVeque}, {\em A stability analysis of neural networks and its application to tsunami early warning}, Journal of Geophysical Research: Machine Learning and Computation, 1 (2024), p.~e2024JH000223, \href{http://dx.doi.org/10.1029/2024JH000223}{doi:\nolinkurl{10.1029/2024JH000223}}.

\bibitem{ross2018improving}
{\sc A.~S. Ross and F.~Doshi-Velez}, {\em Improving the adversarial robustness and interpretability of deep neural networks by regularizing their input gradients}, in AAAI Conference on Artificial Intelligence, 2018.

\bibitem{shafahi2019free}
{\sc A.~Shafahi, M.~Najibi, A.~Ghiasi, Z.~Xu, J.~Dickerson, C.~Studer, L.~S. Davis, G.~Taylor, and T.~Goldstein}, {\em Adversarial training for free!}, in Advances in Neural Information Processing Systems (NeurIPS), 2019.

\bibitem{singla2021skew}
{\sc S.~Singla and S.~Feizi}, {\em Skew orthogonal convolutions}, in Proceedings of the 38th International Conference on Machine Learning (ICML), vol.~139 of Proceedings of Machine Learning Research, 2021.

\bibitem{singla2021householder}
{\sc S.~Singla, S.~Singla, and S.~Feizi}, {\em Householder activations for provable robustness against adversarial attacks}, arXiv preprint arXiv:2108.04062,  (2021).

\bibitem{singla2022improved}
{\sc S.~Singla, S.~Singla, and S.~Feizi}, {\em Improved deterministic $\ell_2$ robustness on {CIFAR-10} and {CIFAR-100}}, in International Conference on Learning Representations (ICLR), 2022.

\bibitem{szegedy2014intriguing}
{\sc C.~Szegedy, W.~Zaremba, I.~Sutskever, J.~Bruna, D.~Erhan, I.~Goodfellow, and R.~Fergus}, {\em Intriguing properties of neural networks}, in International Conference on Learning Representations (ICLR), 2014.

\bibitem{tramer2020adaptive}
{\sc F.~Tram{\`e}r, N.~Carlini, W.~Brendel, and A.~Madry}, {\em On adaptive attacks to adversarial example defenses}, in Advances in Neural Information Processing Systems (NeurIPS), 2020.

\bibitem{tramer2017space}
{\sc F.~Tram{\`e}r, N.~Papernot, I.~Goodfellow, D.~Boneh, and P.~McDaniel}, {\em The space of transferable adversarial examples}, arXiv preprint arXiv:1704.03453,  (2017).

\bibitem{trockman2021orthogonalizing}
{\sc A.~Trockman and J.~Z. Kolter}, {\em Orthogonalizing convolutional layers with the {Cayley} transform}, in International Conference on Learning Representations (ICLR), 2021.

\bibitem{tsipras2019robustness}
{\sc D.~Tsipras, S.~Santurkar, L.~Engstrom, A.~Turner, and A.~Madry}, {\em Robustness may be at odds with accuracy}, in International Conference on Learning Representations (ICLR), 2019.

\bibitem{vaswani2017attention}
{\sc A.~Vaswani, N.~Shazeer, N.~Parmar, J.~Uszkoreit, L.~Jones, A.~N. Gomez, {\L}.~Kaiser, and I.~Polosukhin}, {\em Attention is all you need}, in Advances in Neural Information Processing Systems (NeurIPS), vol.~30, 2017, pp.~5998--6008.

\bibitem{virmaux2018lipschitz}
{\sc A.~Virmaux and K.~Scaman}, {\em {Lipschitz} regularity of deep neural networks: Analysis and efficient estimation}, in Advances in Neural Information Processing Systems (NeurIPS), 2018.

\bibitem{wong2020fast}
{\sc E.~Wong, L.~Rice, and J.~Z. Kolter}, {\em Fast is better than free: Revisiting adversarial training}, in International Conference on Learning Representations (ICLR), 2020.

\bibitem{zhang2019you}
{\sc D.~Zhang, T.~Zhang, Y.~Lu, Z.~Zhu, and B.~Dong}, {\em You only propagate once: Accelerating adversarial training via maximal principle}, in Advances in Neural Information Processing Systems (NeurIPS), 2019.

\end{thebibliography}

\end{document}